\documentclass[10pt,twocolumn,letterpaper]{article}

\usepackage[pagenumbers]{cvpr} 

\usepackage[dvipsnames]{xcolor}  

\newcommand{\david}[1]{{\textcolor{Emerald}{[\textbf{David}: #1]}}}

\usepackage{colortbl}
\colorlet{colorFst}{Green!25}       
\colorlet{colorSnd}{SpringGreen!45} 
\colorlet{colorTrd}{Yellow!30}      
\colorlet{colorLow}{darkgray!30}    
\newcommand{\st}{\cellcolor{colorFst}\bf}   
\newcommand{\nd}{\cellcolor{colorSnd}}      

\usepackage{verbatim}
\usepackage{arydshln}
\usepackage[table]{xcolor}
\usepackage{multirow} 
\usepackage{makecell}

\usepackage[accsupp]{axessibility}  

\usepackage{pifont}
\newcommand{\cmark}{\textcolor{red}{\ding{51}}}%
\newcommand{\xmark}{\textcolor{OliveGreen}{\ding{55}}}%
\usepackage{titletoc}



\definecolor{cvprblue}{rgb}{0.21,0.49,0.74}
\usepackage[pagebackref,breaklinks,colorlinks,allcolors=cvprblue]{hyperref}

\def\paperID{} 
\def\confName{CVPR}
\def\confYear{2026}
\usepackage{balance}
\title{PAGaS: Pixel-Aligned 1DoF Gaussian Splatting for Depth Refinement}

\author{David Recasens\textsuperscript{\rm 1}~~~Robert Maier\textsuperscript{}~~~Aljaz Bozic\textsuperscript{}~~~Stephane Grabli\textsuperscript{}\\Javier Civera\textsuperscript{\rm 1}~~~Tony Tung\textsuperscript{}~~~Edmond Boyer\textsuperscript{}\\\textsuperscript{\rm 1}University of Zaragoza
}

\begin{document}
 \twocolumn[{%
 \renewcommand\twocolumn[1][]{#1}%
 \maketitle
 \centering\includegraphics[width=.87\linewidth]{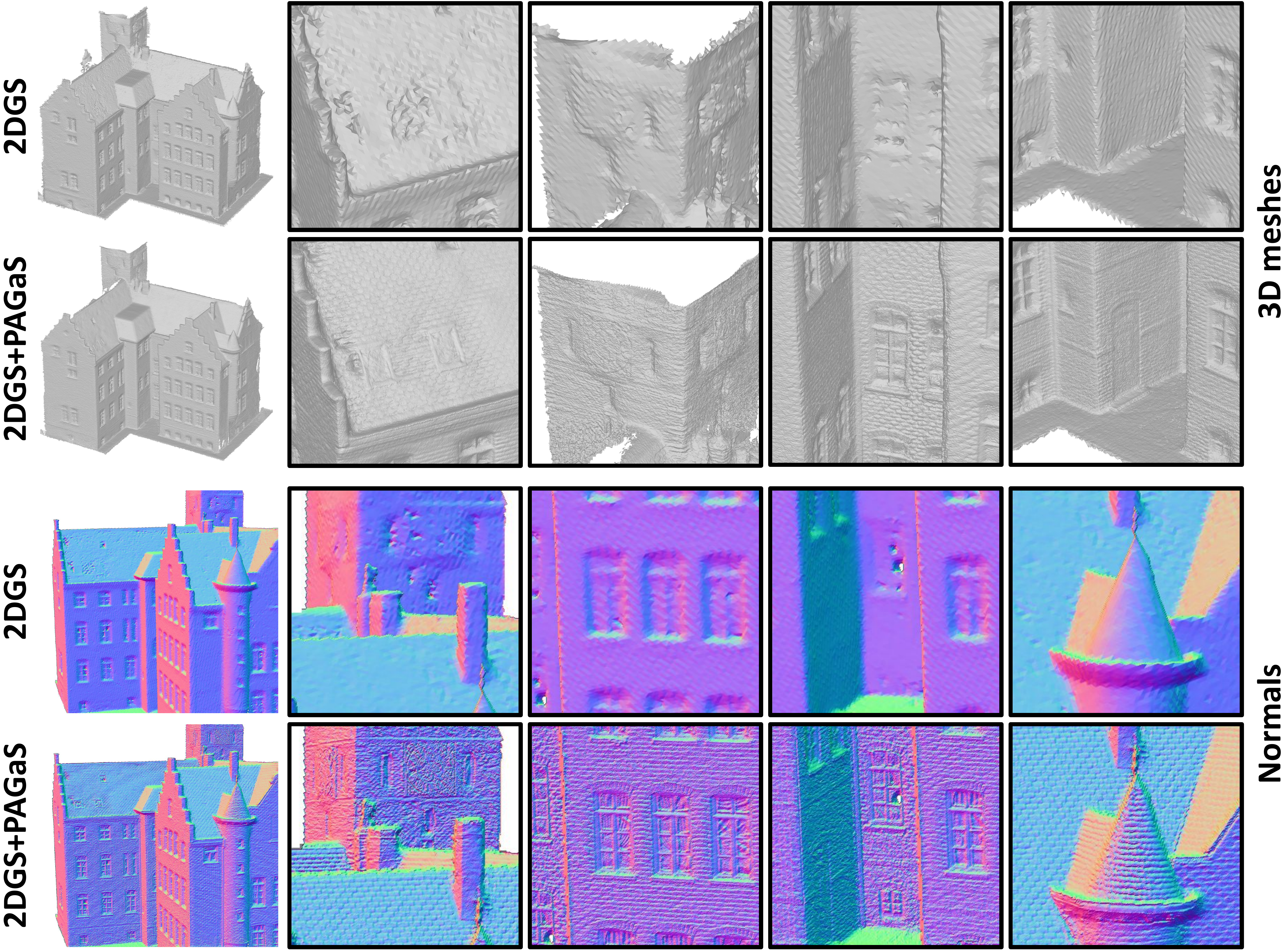}
 \captionof{figure}{
 \textbf{Results for 2DGS and after applying our PAGaS in DTU}. 
 Note the fine-grained details in our refined meshes and depth normals, particularly in the roof tiles, windows, curtains and brick walls, which prior methods fail to capture with this level of fidelity.
 } 
 \vspace{1.45em}
 \label{fig:teaser}
 }
]
 
\begin{abstract}
Gaussian Splatting (GS) has emerged as an efficient approach for high-quality novel view synthesis. While early GS variants struggled to accurately model the scene's geometry, recent advancements constraining the Gaussians' spread and shapes, such as 2D Gaussian Splatting, have significantly improved geometric fidelity. In this paper, we present Pixel-Aligned 1DoF Gaussian Splatting (PAGaS) that adapts the GS representation from novel view synthesis to the multi-view stereo depth task. 
Our key contribution is modeling a pixel's depth using one-degree-of-freedom (1DoF) Gaussians that remain tightly constrained during optimization. Unlike existing approaches, our Gaussians' positions and sizes are restricted by the back-projected pixel volumes, leaving depth as the sole degree of freedom to optimize. PAGaS produces highly detailed depths, as illustrated in \cref{fig:teaser}. 
We quantitatively validate these improvements on top of reference geometric and learning-based multi-view stereo baselines on challenging 3D reconstruction benchmarks. 
Code: \href{https://davidrecasens.github.io/pagas/}{\texttt{davidrecasens.github.io/pagas}}
\end{abstract} 

\section{Introduction}
\label{sec:intro}


Gaussian Splatting (GS)~\cite{kerbl20233d} originally emerged as a highly efficient alternative to Neural Radiance Fields (NeRFs)~\cite{mildenhall2021nerf} for novel view synthesis, with negligible, if any, loss in rendering quality. Due to its adaptive and explicit representation of 3D surfaces, GS has also gained traction as a 3D scene representation~\cite{fei20243d}. Notable examples include offline multi-view reconstructions of static~\cite{guedon2023sugar,zou2024triplane} and dynamic~\cite{yang2024deformable,lin2024gaussian} scenes, editing~\cite{ye2024gaussian,choi2024click}, semantic reconstructions~\cite{qin2024langsplat,shi2024language}, robot navigation~\cite{chen2024splat} and SLAM~\cite{tosi2024nerfs}.

In this paper, we leverage the inherent efficiency of GS for the multi-view stereo depth task, instead of its original goal of novel view synthesis. We are motivated by use cases with a high numbers of views, for which the state of the art, based on visual transformers, does not scale well~\cite{cao2024mvsformer++,izquierdo2025mvsanywhere}. However, while current GS approaches excel at rendering quality at high frame rates, accurately retrieving from them surface geometry remains a research challenge. When GS is optimized using only photometric losses, it typically overfits by overpopulating the 3D space with Gaussians of different transparency levels~\cite{martins2024feature}. In addition, unless they are numerous and very small, the elliptical Gaussian shapes may not be a good fit for generic surface shapes~\cite{guedon2023sugar}.

Similarly to \cite{chen2024pgsr}, we posit that the primary reason for GS overfitting is an excessive amount of optimizable degrees of freedom. Based on that, we propose a radically minimalist novel GS model for multi-view stereo depth, that we denote as Pixel-Aligned 1DoF Gaussian Splatting and shorten as PAGaS. Our main contribution is the use of per-pixel spherical Gaussians for fitting back-projected pixel volumes. In contrast to general GS approaches in the literature, our Gaussians are fixed in number (the number of image pixels) and have their depth along the ray as the only optimizable parameter. We handcraft their shape to be spherical and their radius to fit the back-projected pixels, so neither their size nor their orientation is estimated. 

The standard procedure to extract a 3D mesh from a GS model starts by alpha-blending Gaussians to obtain the per-view depth maps~\cite{huang20242dgs,guedon2023sugar,chen2024pgsr}, and then fusing them using a Truncated Signed Distance Function (TSDF)~\cite{curless1996volumetric}. As 3D reconstruction inevitably involves rendering  pixel-wise depth, we reconfigure the optimization strategy to obtain the best possible depth maps. As geometric accuracy improves with the number of Gaussians~\cite{fang2024mini}, scaling it to the minimum size of one pixel is reasonable.  As optimizing pixel-wise Gaussians for all views at the same time would have high memory requirements and overlaps in the 3D, we optimize the pixel-wise Gaussians of each view independently and sequentially.

As a result, our PAGaS is a very effective model for refining an existing depth map into one with much higher detail using photometric constraints from multiple overlapping views. 
Thanks to its efficiency, it requires only a few seconds per frame and relatively small GPU memory, even with high resolution images. It can be applied as post-processing to any 3D reconstruction method, significantly improving fine-grained structural details as shown in \cref{fig:teaser}.
We provide additional qualitative and quantitative results, showing how our PAGaS can refine depths coming from distinct methods and in multiple public datasets.

\section{Related Work}
\label{sec:related}


\subsection{Neural Reconstructions}

While showing impressive image synthesis results, volume rendering approaches~\cite{mildenhall2021nerf} often fail to produce high-quality surface reconstructions~\cite{remondino2023critical}, as photometric losses do not directly enforce it and MLPs have a smoothness inductive bias. Direct encoding of geometric biases is convenient for enhancing geometric accuracy. For example, NeuS~\cite{wang2021neus,wang2023neus2} represents surfaces as the zero-level set of an implicit signed distance function (iSDF). 
NeusG~\cite{chen2023neusg} improves the geometry by 3DGS surface regularization, and UniSDF~\cite{wang2023unisdf} deals with reflections but, due to their iSDF backbone, they remain computationally expensive.
ProbeSDF~\cite{toussaint2024probesdf} accelerates optimization by decoupling angular and spatial components in its formulation, while methods such as Sun et al.~\cite{sun2022direct} and Wu et al.~\cite{wu2022voxurf}, use explicit voxel grids to represent geometry. Neuralangelo~\cite{li2023neuralangelo} uses multi-resolution 3D hash grids in combination with neural surface rendering. 
In the same spirit, our PAGaS encodes a multi-view inductive bias by tailoring the shape of the Gaussians to the back-propagated pixels.

\subsection{Splatted Reconstructions}

As in neural reconstructions, the inductive bias of 3DGS representations~\cite{kerbl20233dgs} have a strong influence on the resulting geometry. Specifically, excessive degrees of freedom tend to generate unstructured Gaussian clouds that overfit the photometry. Overfitting has been reduced by increasing the Gaussians' expressivity, and hence reducing their number~\cite{martins2024feature}. 2DGS~\cite{huang20242dgs}, SuGaR~\cite{guedon2023sugar}, PGSR~\cite{chen2024pgsr}, the 3DGS post-processor~\cite{zhangeve3d},
and all 3DGS reconstruction methods ~\cite{chen2024vcr,turkulainen2025dn,zhang2024quadratic}, 
optimize a single cloud of Gaussians that move freely in three-dimensional space while fitting all camera views jointly, regularizing their normals to align with the surface. In our approach, on the other hand, each input image has its own cloud of pixel-aligned Gaussians, where each Gaussian is allowed to slide only along its back-projected pixel ray, optimizing one view at a time. In addition, our Gaussians' depth are the only optimizable parameter, contrasting with the high number of variables of all other methods. 
RaDe-GS~\cite{zhang2024radegs} focuses on the problem of rasterizing a depth map from an existing GS representation. In contrast, we formulate the depth optimization problem with a GS representation.

\subsection{Data-driven Reconstructions}

While most GS approaches optimize a representation, recent alternatives have replaced parts of the standard GS pipeline by learned models. For example, PixelSplat~\cite{charatan2024pixelsplat} can learn the Gaussian parameters from data and infer them from two views. MVSplat~\cite{chen2024mvsplat} predicts separately the means of the Gaussians with a deep neural encoder. Splatter Image~\cite{szymanowicz2024splatter} operates by predicting pixel-wise Gaussians from a single view. DepthSplat~\cite{xu2024depthsplat} connects single-view depth networks and Gaussian Splatting representations for consistency. These networks achieve competitive novel-view color synthesis results, but struggle in capturing consistent 3D reconstructions. 
Models like Vis-MVSNet~\cite{zhang2023vis}, RobustMVD~\cite{schroppel2022benchmark}, MVSFormer++~\cite{cao2024mvsformer++}, MVSAnywhere~\cite{izquierdo2025mvsanywhere} and similar~\cite{cao2022mvsformer,yu2020fast,yang2022mvs2d,wang2021patchmatchnet}, dedicated solely to multi-view stereo depth estimation~\cite{wang2024learning}, still outperform them in accuracy. 
In comparison, optimization-based approaches such as ours yield more consistent and sharper reconstructions by using raw pixel colors at test time instead of relying on learned patterns.

\section{Method}
\label{sec:method}

To obtain a single, globally consistent 3D reconstruction from multiple 2D images, one that preserves the fidelity of the input views without hallucinating content, it is essential to account for information from all viewpoints. However, global optimization methods designed to enforce cross-view consistency often struggle to recover fine-grained details and typically scale poorly with increasing image count or resolution. 
In contrast, high-frequency depth refinement is inherently local and should be addressed at the pixel level. While enforcing some degree of neighborhood consistency is beneficial, enforcing full scene-level global consistency greatly increases computational cost without significant gains in accuracy.

Building on this idea, PAGaS sequentially refines the depth map of each input image, starting from the coarse but globally consistent geometry produced by a baseline method (e.g., 2DGS). This allows us to inject high-frequency details into the reconstruction without altering its overall structural coherence. Our goal is to recover the finest geometric details encoded in the input images by repurposing 3DGS as a lightweight optimization framework for the multivew stereo task.

\begin{figure}[ht!]
  \centering
  \includegraphics[width=1.\linewidth]{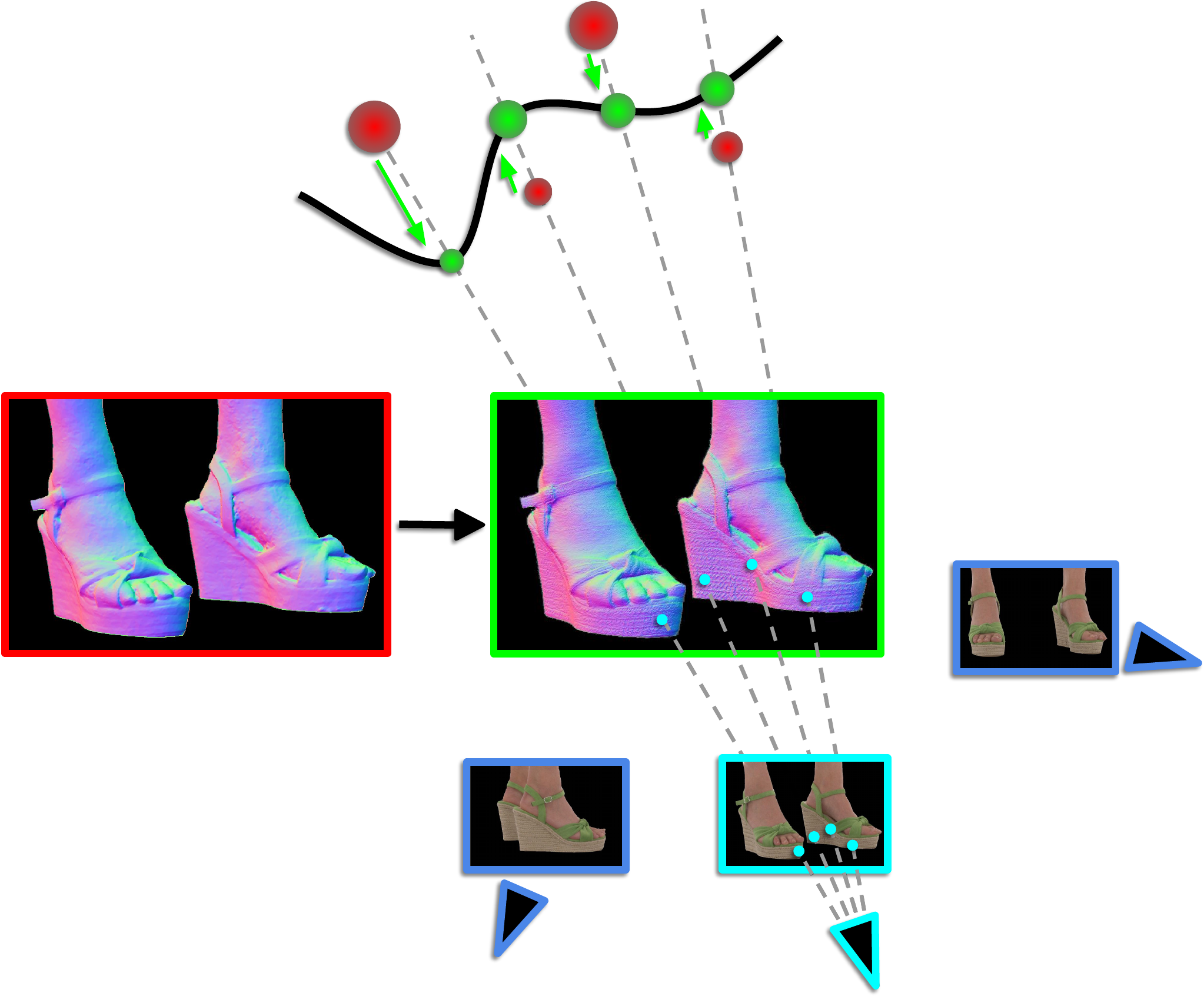}
  \caption{\textbf{Depth refinement with Pixel-Aligned Gaussian Splatting}. For each valid pixel of the \textcolor[RGB]{0,230,230}{target view} to refine, a 3D Gaussian is initialized along its camera-to-pixel ray, at the position determined by its \textcolor[RGB]{255,64,64}{initial coarse depth}. Gaussians are spherical (trivializing the rotation) with a depth dependent-scale (the further the bigger). Gaussian appearance is locked to its pixel color value, and opacity fixed to 1. \textcolor[RGB]{96,128,192}{Stereo views} are only used to constrain the \textcolor[RGB]{100,243,0}{optimization} against the splatted images in a photometric loss.}
  \label{fig:overview}
\end{figure}

\subsection{Preliminaries}

\paragraph{3D Gaussian Splatting.} 
3DGS explicitly represents 3D scenes using a set of $M$ 3D Gaussian primitives 
defined by their 3D position $\boldsymbol{\mu} \in \mathbb{R}^3$ and covariance $\boldsymbol{\Sigma} \in \mathbb{S}_+^{3}$

\begin{equation}
   G(\mathbf{p}) = e^{-(\mathbf{p} - \boldsymbol{\mu})^T \boldsymbol{\Sigma}^{-1}(\mathbf{p} - \boldsymbol{\mu})} 
\label{eq:1}
\end{equation}

The covariance $\boldsymbol{\Sigma}$ is defined by a scale $\mathbf{s} \in \mathbb{R}^{3}$ and a rotation matrix $\mathbf{R} \in SO(3)$, which together define the orientation and anisotropy of the 3D Gaussian through $\boldsymbol{\Sigma} = \mathbf{R} \mathbf{S} \mathbf{S}^T \mathbf{R}^T$. $\mathbf{S}$ stands for the diagonal matrix formed with the scaling vector $\mathbf{s}$. Gaussians are projected to the image plane via local affine transformation \cite{zwicker2001ewa} $\mathbf{J}$, yielding $\boldsymbol{\Sigma}' = \mathbf{J}\mathbf{W}\boldsymbol{\Sigma}\mathbf{W}^T\mathbf{J}^T$. The corresponding 2D Gaussian $G^{2D}$ is obtained by discarding the third column of $\boldsymbol{\Sigma}'$.
2D Gaussians are then sorted by their depth at each pixel $\mathbf{u}$, and color $c$ by alpha-blending

\begin{equation}
   c({\mathbf{u}}) = \sum_{k=1}^M c_k \mathbf{\alpha}_k G^{2D}({\mathbf{u}})            \prod_{j=1}^{k-1} (1 - \alpha_jG^{2D}({\mathbf{u}}))
\label{eq:2}
\end{equation}

\noindent where the visibility of each primitive $k$ is controlled by its opacity $\mathbf{\alpha}_k \in \mathbb{R}$, and view-dependent appearance $c_k$. 

\paragraph{Multi-View Depth Map Refinement.} 

We focus on refining all the valid pixel values of an input depth map $D$ at a given target view, using $N$ neighboring context views only for photometric stereo constraints during the optimization. We assume known camera extrinsics and intrinsics.
We adopt a simple strategy to select the $N$ context views based on the dot product of the respective camera viewing directions. In datasets with different kinds of cameras, we group them by similar field-of-views, in order to prevent aliasing.

\subsection{Pixel-Aligned Gaussians for Direct Depth Optimization}

Prior 3DGS-based geometry reconstruction methods optimize a single global set of Gaussians and recover the scene’s 3D mesh by splatting them across all views to render depth maps, which are then fused into a TSDF.
We take a fundamentally different approach: instead of optimizing high–degrees-of-freedom 3D Gaussians in world space, we reformulate reconstruction as a purely 2D optimization over depth maps parameterized by constrained 1DoF Gaussians.
In standard 3DGS pipelines, Gaussians can move freely in 3D to simultaneously satisfy all views. 
In contrast, we optimize one view at a time, assigning a single Gaussian per pixel and restricting its motion strictly along the pixel’s back-projected camera ray. This design allows us to exploit pixel-level information directly while preserving global consistency through multi-view cues. 
As illustrated in \cref{fig:overview}, Gaussians are instantiated only in the target view, while depth convergence is guided by the $N$ context stereo views. The refined depth for each pixel is simply the optimized $z$ value of its pixel-aligned Gaussian.

\subsection{1DoF Gaussian Parameters Conditioning}

At the core of our method, we use the concept of 1DoF Gaussians as 3D pixel-aligned Gaussians whose only optimizable parameter is their depth. We condition all per-Gaussian parameters ($\boldsymbol{\mu}$, $\mathbf{\alpha}_k$, $\mathbf{s}$, $\mathbf{R}$) on this depth instead of optimizing them independently.
The 3D positions of the Gaussians are directly computed from the depth map by back-projecting to 3D vertex positions using the intrinsic camera parameters.
Instead of using the Gaussians' rotations, we directly derive pixel normals from the 3D vertex map through a cross product of the neighboring pixels around the current one using central differences.
As we design Gaussians to be pixel-aligned, we adopt isotropic spherical Gaussians ($\mathbf{s} = s\mathbf{I}$).
Since we have a Gaussian per pixel, each Gaussian should have a scale that always fills its target view pixel, minimizing overlap with adjacent Gaussians to avoid perturbing their rendered pixel colors.
A co-optimization of separate pixel-wise scales per-Gaussian would prevent optimization of pixel-wise Gaussians, and instead favor larger individual Gaussians that cover more than a pixel. Therefore, we analytically set it to always cover its target view pixel, independently of their depth value. 
To achieve Gaussians with the size of 1 pixel in image space, we scale the Gaussians linearly with their Euclidean depth $d_e$, determined by the depth $d$ along the target view pixel rays and camera intrinsics ($f_x$, $f_y$, $c_x$, $c_y$)

\begin{equation}
   d_e = d \left({{\left(\frac{u - c_x}{f_x}\right)^2 + \left(\frac{v - c_y}{f_y}\right)^2 + 1}}\right)^{-\frac{1}{2}}
\label{eq:3}
\end{equation}

\noindent setting the scale as approximately half of the side of a pixel back-projection at depth $\mathbf{d}_e$

\begin{equation}
   s = \frac{d_e }{2\sqrt{f_x  f_y} }
\label{eq:4}
\end{equation}


We sample the pixel colors associated to the target view pixel and use them to initialize the Gaussian colors, represented as zero-order spherical harmonics. Fixing the appearance during optimization removes additional degrees of freedom from the optimization by preventing view-dependent effects.

Since we assume spherical 3D Gaussians with constant appearance, we set the per-Gaussian rotation to identity without impact on rasterization.

Instead of optimizing the opacities of the Gaussians, we follow the intuition that each pixel has a surface at the pixel's depth and hence set the opacity to fully opaque.
By preventing mixing 3D structure with view-dependent colors and opacities, Gaussians are forced to move along the ray to the correct position.

Compared to the original 3DGS formulation, we dramatically reduce the number of optimization parameters from 59 to just one (depth), with the original parameters being: 48 (3 RGB channels $\times$ 16) for 3rd-order spherical harmonics color, 3 for means, 4 for rotation (quaternions), 3 for scale, and 1 for opacity~\cite{bagdasarian20243dgs}.

\begin{figure}[ht!]
  \centering
  \includegraphics[width=1.\linewidth]{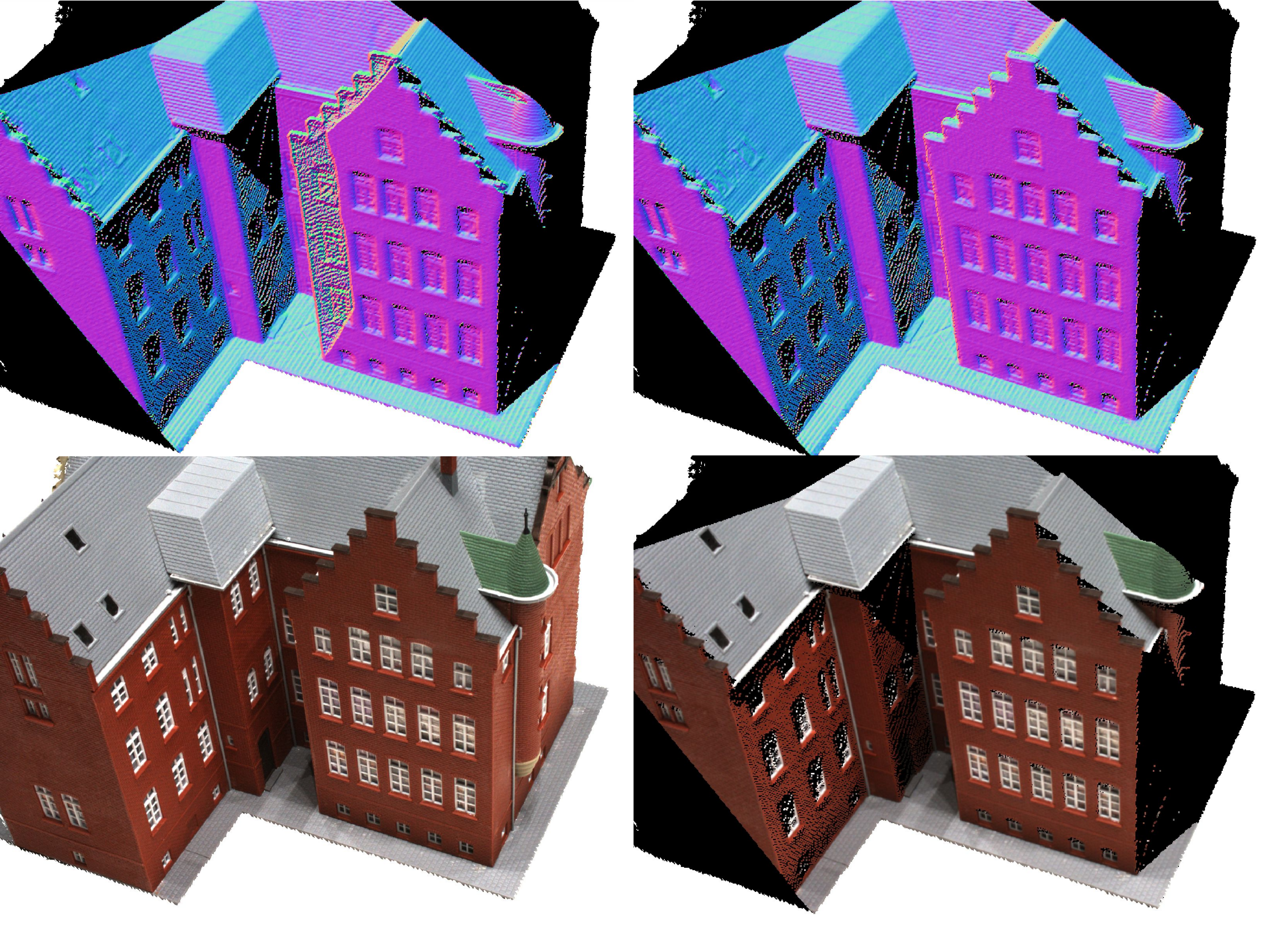}
  \caption{\textbf{Occlusions problem (top)} caused by having Gaussians only from the target view. 
  These images show the normals from the rendered depth at a context view without and with the radius and depth thresholds of the Occlusion-Aware 3DGS Rasterizer. See how the Gaussians behind the wall perturbe its rendered values. In addition, some areas can have poor Gaussian coverage, provoking that those Gaussians that should be occluded are actually rendered in the target view. Increasing the radius threshold allows the sparse Gaussians of the front to cover a wider area and block them. \textbf{Disocclusions problem (bottom)}. We warp target-view pixels using their initial depths to mask out regions in the context views that are not visible from the target view. Left and right images show a context view before and after applying this mask (black pixels). By comparing the masked color context view with the optimized rendered normal from depth above, we can see how the disoccluded (black) areas align.}
  \label{fig:occlusions}
\end{figure}

\begin{figure}
  \centering
  \vspace*{-\baselineskip}  
  \includegraphics[width=\linewidth]{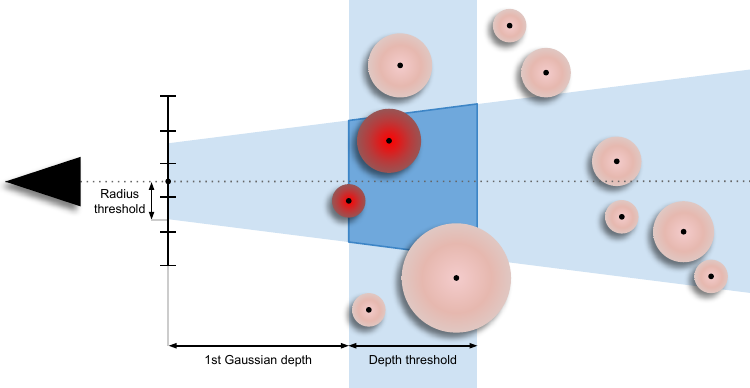}
  \caption{\textbf{Occlusion-Aware 3DGS Rasterizer} logic, newly proposed in this paper to effectively ignore Gaussians that should be occluded during alpha-blending. \textbf{Radius threshold} delimits in pixel units a circular area, around the center of the pixel that is being rasterized, where only Gaussians with a 2D mean that falls inside it are considered during alpha-blending. \textbf{Depth threshold} is a depth value that is added to the depth of the first Gaussian whose 2D mean falls inside the radius threshold. Gaussians farther than this depth limit are ignored in the rasterization. Both thresholds define a truncated cone volume in the 3D (darker blue area) of valid (darker) Gaussians when rendering each pixel.}
  \label{fig:rasterizer}
\end{figure}

\subsection{Occlusion-Aware 3DGS Rasterizer}

When rendering the Gaussians of the target view that is being optimized from the context viewpoints used for supervision, occlusions can arise, causing some Gaussians to overlap during rasterization. Due to the pixel-wise and ray-aligned nature of our approach, we do not dynamically create or reposition Gaussians to increase coverage. Consequently, both foreground Gaussians and those that should be occluded are included in the alpha-blending sum of pixels with overlap of Gaussians. Top-left image in \cref{fig:occlusions} illustrates this problem in the rendered normal from depth at pixels with Gaussians that should be ignored. However, Gaussians of nearby pixels that belong to the same surface can overlap when rendering from other views, only the closest ones should be included in the rasterization.
In order to only occlude these Gaussians that do not belong to the same surface, we implemented two new hyperparameters in the alpha-blending logic of 3DGS, showcased in \cref{fig:rasterizer}: \emph{depth} and \emph{radius thresholds}.

\paragraph{The radius threshold} establishes a hard limit on the maximum distance in pixel space at which the mean of a splatted Gaussian can be included in the alpha blending of that pixel. We analytically calculate the scale of the spherical Gaussians to constrain their influence around their corresponding pixels in the target view. However, Gaussians in the context view’s screen space do not necessarily cover only one pixel; they can be larger. With this radius, we ensure that Gaussians maintain a limited range of influence across all views.

\paragraph{The depth threshold} effectively masks out Gaussians that fall within the radius threshold of the pixel being rasterized but are located far behind the front ones. Since Gaussians are added during alpha blending in a sort manner (i.e., the closest ones to the camera first), we add an input depth threshold to the depth of the first Gaussian that falls within the radius of influence. This delimits the maximum depth a Gaussian can have to be included in the rasterization. Depth threshold value is automatically defined by the depth range of the initialization depth divided by an input number of slices.

We implement this new logic in CUDA on top of the Gsplat~\cite{ye2024gsplatopensourcelibrarygaussian} 3DGS rasterizer.

\subsection{Loss functions}

\paragraph{Photometric Consistency Loss.}
In the multi-view photometric consistency loss, we minimize the $L_1$ loss and $SSIM$ of the rendered color images at target and context views with their corresponding input color images, with a hyperparameter weight $\lambda_c$

\begin{equation}
   \mathcal{L}_c = L_1 (1-\lambda_c) + \lambda_c SSIM
\label{eq:5}
\end{equation}

Pixel-wise Gaussians, by nature, cover only the surface visible in the target view, while context views have a different perspective and can reveal areas where Gaussians do not fully cover the surface. This produces an incorrect color gradient in these lonely Gaussians due to blending with the default black background.
This is a common problem when a surface is nearly parallel to the viewing rays of the target view. To mitigate this problem, in the $L_1$ loss we approximately mask out the pixels of the context views that capture these low density areas. This is done by warping the target valid pixels over all context views using the initial depth values, and removing all pixels where a target view pixel does not fall. The bottom row of \cref{fig:occlusions} illustrates the original and the masked out context view. 

We further incorporate 
a per-pixel color gradient weight $(w_{grad})$ into the $L_1$ loss function,
minimizing the influence of pixels with low color gradient relative to their neighbors ($c_{i}$, $c_{j}$), normalizing within specific limits $grad_{min}$, $grad_{max}$

\begin{equation}
w_{grad}= \frac{|c_{i} - c_{j}| - grad_{min}}{grad_{max} - grad_{min}} 
\label{eq:6}
\end{equation}

\paragraph{Normal Smoothness Loss.}
To minimize the aleatoric uncertainty of the input pixel colors, we add a normal smoothness loss which maximizes the dot product between adjacent pixel normals ($\mathbf{n}_i$, $\mathbf{n}_j$) --extracted from the optimized depth map with central differences. In the target view, this is weighted by a factor $\lambda_s$ and the opposite of the $w_{grad}$ to smooth only low color gradient pixels

\begin{equation}
   \mathcal{L}_s = \frac{\lambda_s}{2}\left( 1 - \sum \mathbf{n}_i \mathbf{n}_j     \right)(1 - w_{grad})
\label{eq:7}
\end{equation}

The final loss being $\mathcal{L} = \mathcal{L}_c + \mathcal{L}_s$.

When foreground masks are available (e.g., from segmentation networks) we use these to optimize only the Gaussians of foreground pixels within the mask.

We embed our optimization in a coarse-to-fine resolution pyramid scheme, where we first optimize at lower, downsampled resolutions and progressively refine the depth maps up to the original high resolution. This strategy accelerates convergence and suppresses noise in the input depth maps introduced by the initial downsampling. After optimizing at each pyramid level, we linearly upsample the refined depth map to initialize the next finer level.
\section{Experiments}
\label{sec:experiments}

\begin{table}[t]
    \centering
    \resizebox{\columnwidth}{!}{%
    \begin{tabular}{lccccccc}
        \toprule
        \multicolumn{1}{c}{} &
        \multirowcell{2}[-1mm]{\textbf{Stereo}\\\textbf{views}} &
        \multirowcell{2}[-1mm]{\textbf{Pretrain}} &
        \multirowcell{2}[-1mm]{\textbf{SfM}\\\textbf{initialization}} &
        \multicolumn{2}{c}{\textbf{Resolution}} &
        \multicolumn{2}{c}{\textbf{Time / Frame (s)}} \\
        \cmidrule(lr){5-6} \cmidrule(lr){7-8}
        & & & & \textbf{DTU} & \textbf{TNT} & \textbf{DTU} & \textbf{TNT} \\
        \midrule
        MVSAnywhere & 7   & \cmark & \xmark & 640$\times$480   & 640$\times$480   & 0.12 & 0.12 \\
        2DGS        & All & \xmark & \cmark & 777$\times$581   & 978$\times$546   & 24   & 8 \\
        PGSR        & All & \xmark & \cmark & 777$\times$581   & 978$\times$546   & 37   & 11 \\
        PAGaS       & 10  & \xmark & \xmark & 1554$\times$1162 & 1954$\times$1090 & 10   & 11 \\
        \bottomrule
    \end{tabular}
    }
    \caption{Comparison of the different multi-view stereo depth estimation methods on number of used stereo views, if using pretrained networks, Structure-from-Motion initialization~\cite{schonberger2016structure}, input image resolution, and time per-frame. 2DGS and PGSR time is averaged for the number of views in scenes Scan24 for DTU and Truck for TnT. MVSAnywhere, 2DGS and PAGaS fit in a single 11GB VRAM graphic card. PGSR needs a 24GB card. MVSAnywhere estimation resolution is fixed to 640x480 due to its transformer architecture. 2DGS and PGSR optimization time and VRAM requirements increase considerably if not half resolution --4 times less input information-- is used. PAGaS is the only method that leverages full-resolution input images to estimate highly detailed depth at the same size, without pretraining, without SfM point cloud initialization, excessive GPU memory, and within a reasonable runtime.}
    \label{tab:methods_comparison}
\end{table}

\begin{table*}[t]
    \centering
    \footnotesize
    \setlength{\tabcolsep}{4pt}
    \resizebox{\textwidth}{!}{%
    \begin{tabular}{@{} l *{15}{c} c @{}}
        \toprule
         & \textbf{24} & \textbf{37} & \textbf{40} & \textbf{55} & \textbf{63} & \textbf{65} & \textbf{69} & \textbf{83} & \textbf{97} & \textbf{105} & \textbf{106} & \textbf{110} & \textbf{114} & \textbf{118} & \textbf{122} & \textbf{Mean} \\
        \midrule
        MVSAnywhere     & 2.08 & 1.90 & 1.19 & 1.33 & \nd2.15 & \nd1.92 & 1.99 & 2.42 & 2.38 & 1.43 & 2.28 & 2.74 & 1.35 & 2.46 & 1.94 & 1.97 \\
        MVSAnywhere + PAGaS        & \nd 1.80 & \nd1.88 & \nd0.98 & \nd1.04 & 3.04 & \nd1.92 &\nd 1.84 & \nd2.28 & \nd2.16 & \nd1.22 & \nd2.09 & \nd2.38 & \nd1.14 & \nd2.28 & \nd1.67 & \st1.85 \\
        \noalign{\vskip 1mm} 
        \hdashline 
        \noalign{\vskip 1mm}
        2DGS            & 0.50 & 0.79 & 0.34 & \nd0.44 & \nd0.81 & \nd0.87 & 0.89 & 1.20 & 1.17 & 0.63 & \nd0.69 & 1.29 & 0.47 & \nd0.68 & \nd0.47 & 0.75 \\
        2DGS + PAGaS               & \nd0.46 & \nd0.72 & \nd0.32 & 0.45 & \nd0.81 & 0.89 & \nd0.83 & \nd1.08 & \nd1.13 & \nd0.58 & \nd0.69 & \nd1.21 & \nd0.44 & 0.70 & 0.49 &\st0.72 \\
        \noalign{\vskip 1mm} 
        \hdashline 
        \noalign{\vskip 1mm}
        PGSR            & \nd0.37 & 0.55 & 0.41 & \nd0.34 & 0.78 & \nd0.57 & \nd0.49 & 1.08 & 0.86 & 0.58 & \nd0.48 & 0.49 & \nd0.31 & \nd0.37 & \nd0.34 & 0.54 \\
        PGSR + PAGaS              & 0.39 & \nd0.48 & \nd0.34 & 0.36 & \nd0.69 & 0.60 & \nd0.49 & \nd0.94 & \nd0.84 & \nd0.52 & 0.49 & \nd0.47 & \nd0.31 & 0.40 & 0.36 & \st0.51 \\
        \bottomrule
    \end{tabular}}
    \caption{\textbf{Quantitative DTU evaluation} showing Chamfer distance in mm $\downarrow$ for baseline methods before and after our PAGaS refinement.} 
    \label{tab:metrics_dtu}
\end{table*}

\begin{table*}[t]
    \centering
    \footnotesize
    \setlength{\tabcolsep}{4pt} 
    \resizebox{.8\textwidth}{!}{%
    \begin{tabular}{lccccccc}
        \toprule
        \textbf{Scene} & \textbf{Barn} & \textbf{Caterpillar} & \textbf{Courthouse} & \textbf{Ignatius} & \textbf{Meetingroom} & \textbf{Truck} & \textbf{Mean} \\
        \midrule        
        MVSAnywhere         & 0.21 & 0.11 & 0.08 & 0.22 & \nd0.16 & 0.18 & 0.16 \\
        MVSAnywhere + PAGaS & \nd0.29 & \nd0.12 & \nd0.09   & \nd0.25 & 0.15  & \nd0.21 & \st0.18 \\
        \noalign{\vskip 1mm} \hdashline \noalign{\vskip 1mm}        
        2DGS                & 0.40 & 0.25 & \nd0.17 & 0.51 & \nd0.05 & 0.18 & 0.26 \\
        2DGS + PAGaS        & \nd0.42 & \nd0.27 & \nd0.17 & \nd0.57 & \nd0.05 & \nd0.19 & \st0.28 \\
        \noalign{\vskip 1mm} \hdashline \noalign{\vskip 1mm}
        PGSR                & \nd0.54 & 0.44 & \nd0.24 & 0.80 & 0.13 & \nd0.23 & \st0.40 \\
        PGSR + PAGaS        & 0.52   & \nd0.45   & \nd0.24   & \nd0.81   & \nd0.14   & \nd0.23   & \st0.40   \\
        \bottomrule
    \end{tabular}}
    \caption{\textbf{Quantitative TnT evaluation} showing the F1-score $\uparrow$ for baseline methods before and after our PAGaS refinement.}
    \label{tab:metrics_tnt}
\end{table*}

\begin{figure*}
  \centering
  \includegraphics[width=1.\linewidth]{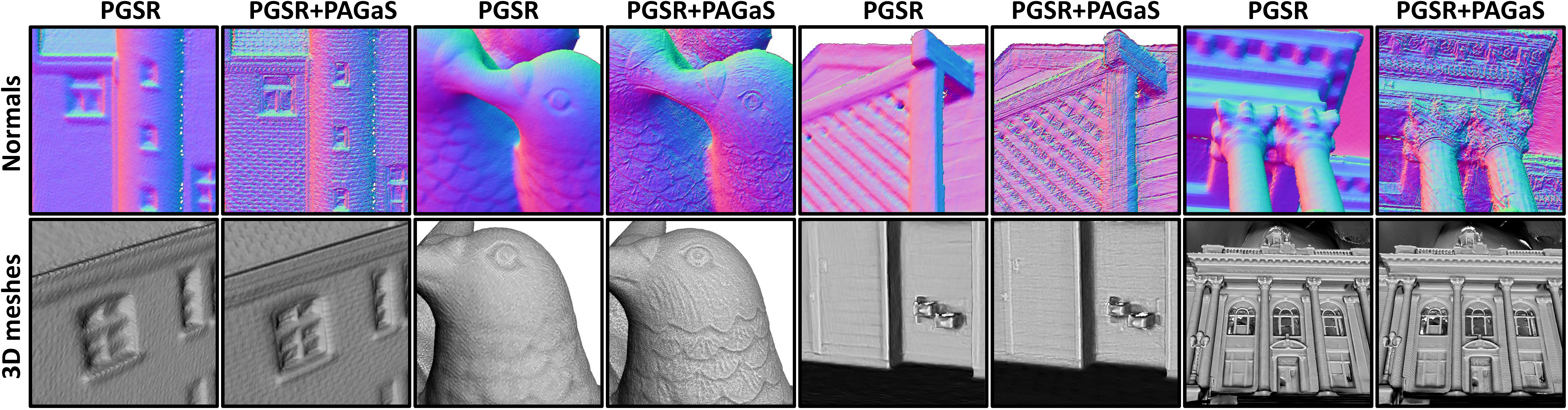}
  \caption{
  \textbf{Normals from depth and 3D meshes from PGSR before and after refinement with PAGaS} on DTU (scan24, scan106) and TnT (Barn, Courthouse). The normal maps reveal how PAGaS recovers the finest pixel-level surface details. All refined depth maps are fused into a single mesh using a TSDF. Under comparable memory budgets, smaller scenes allow finer voxel sizes, which preserve high-frequency geometry more effectively. Consequently, the local depth refinements introduced by PAGaS are clearly visible in compact scenes such as DTU, moderately visible in big-scale scenes like Barn, and largely suppressed in large environments such as Courthouse.
  }
  \label{fig:normals_and_meshes_pgsr}
\end{figure*}

\begin{figure}
  \centering
  \includegraphics[width=1.\linewidth]{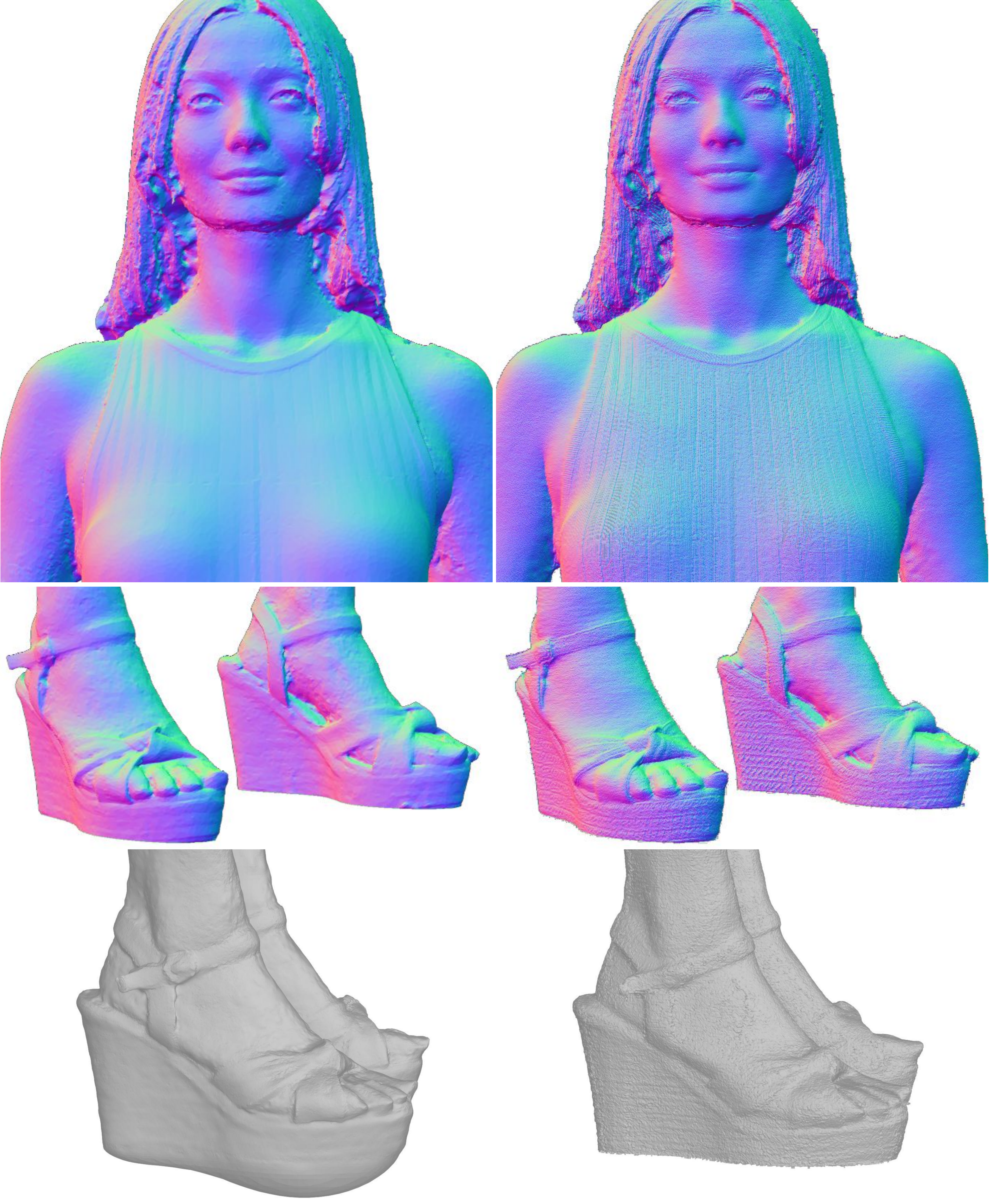}
  \caption{\textbf{ActorsHQ normals from depth and meshes} from Colmap (left) and after PAGaS refinement (right). Zoom in to observe fine-grained details, e.g., at the eyelashes and eyebrows, the fabric of the dress or the sole of the sandals.}
  \label{fig:actor_normal_and_mesh}
\end{figure}

\paragraph{Datasets \& Metrics.} For our evaluation we use 4 different public datasets: DTU, Tanks-and-Temples (TnT), BlendedMVS and ActorsHQ. 
DTU~\cite{jensen2014large} contains varied objects recorded from a robotic arm with movable light sources. 
TnT~\cite{knapitsch2017tanks} records in-the-wild large-scale scenes with inconsistent camera light exposure.
We use these datasets to present our quantitative results, as they are the standard and most widely adopted benchmarks in multi-view 3D reconstruction, thanks to their high quality ground truth.
ActorsHQ~\cite{isik2023humanrf} and BlendedMVS low-res set~\cite{yao2020blendedmvs} are object-centric datasets without ground truth 3D reconstructions that we use for showing additional qualitative results. 
Note that ActorsHQ captures with 160 cameras at high resolution (2990x4088). 
We use this dataset to demonstrate our method’s scalability to both a large number of cameras and high resolution data. The baselines cannot operate at this scale in practice. They require aggressive input downsampling or a reduced view set, which drastically removes high frequency information.

In DTU, we report the Chamfer distance, the bidirectional average nearest-neighbor distance between two point clouds. In TnT, we use the F1-score, the harmonic mean of precision and recall, measuring how many predicted points lie near the ground truth and vice versa.

\paragraph{Baselines.} 
Our quantitatively evaluated multi-view stereo baselines are the widely-used 2DGS~\cite{huang20242dgs}, the state-of-the-art optimization-based PGSR~\cite{chen2024pgsr} and the state-of-the-art learning-based MVSAnywhere~\cite{izquierdo2025mvsanywhere}.
For additional qualitative analysis, we use the gold standard COLMAP~\cite{schoenberger2016mvs}'s multi-view stereo. For each baseline, we compute its full mesh for a multi-view image set using the provided camera poses. We then render individual depth maps from the mesh, refine them using PAGaS, and fuse the refined depths using a TSDF. Finally, we extract a triangle mesh using the Marching Cubes~\cite{lorensen1998marching} algorithm.

\paragraph{Optimization.}
We use two pyramid levels: we first halved the input depth (previously adjusted to the input image resolution) and optimize during 200 iterations (100 in DTU), then upsample to the full resolution and perform an additional 100 iterations. Our method requires only a few iterations at both pyramid levels to converge, as we focus on refining surface details from a given geometry. We stop the Adam optimization of any scale when the learning rate falls below $10^{-7}$, determined by the scheduler that reduces it by 10 when the loss gets stuck for 10 iterations. We start each scale with $10^{-5}$. Our hyperparameters are $\lambda_c=0.2$, $\lambda_c=0.2$, $grad_{min}=0.02$, $grad_{min}=0.1$. Radius threshold as 1.42 pixels in DTU and TnT, and 2 pixels in ActorsHQ and BlendedMVS. Number of slices for depth threshold as 20. TSDF uses a voxel resolution of 1mm in DTU and of 4-6mm in TnT (the latter for the largest Courthouse and Meetingroom scenes).
\cref{tab:methods_comparison} compares PAGaS with the main baselines in terms of the computational cost of per-image depth estimation on a Nvidia A6000.

\subsection{Quantitative Evaluation}
We quantitatively evaluate the effect in the 3D meshes of our depth refiner in \cref{tab:metrics_dtu} and \cref{tab:metrics_tnt} via post-processing state-of-the-art multi-view 3D reconstruction methods, both learning-based (MVSAnywhere) and optimization-based (2DGS and PGSR), on DTU and TnT, and using their publicly available open-source implementations.
PGSR originally preprocess its input point cloud and camera intrinsics and extrinsics using Colmap bundle adjustment, and in TnT also bounds the estimated depth before the TSDF using the ground truth point cloud. For a fair comparison, all our evaluations have been tested under the same input data and ground truth-free depth fusion.
On TnT, PGSR optimizes a per-image illumination exposure model. To ensure a more equitable comparison, we handle the inconsistent illumination there by similarly compensating for exposure and applying a multi-view stereo consistency check~\cite{schonberger2016pixelwise}.

Chamfer distance and F1-score are metrics that primarily evaluate the global structure of a reconstruction, and are less sensitive to fine-scale surface detail. Although our goal is to enrich the input depths and meshes with pixel-level refinements without altering their overall geometry, PAGaS still generally delivers quantitative gains across methods on both datasets, with particularly strong improvements on the small-scale scenes of DTU.
Unlike 2DGS and PGSR, PAGaS does not model view-dependent Gaussian color.
This doubles our inference speed with marginal impact in most cases, suffering in scenes with heavy inconsistent illumination. Notice how PAGaS is sensitive to similar unstable lighting scenes of DTU (scan55, scan65, scan118 and scan122) for both GS-based baselines, but not with respect to MVSAnywhere.

Thanks to its post-processing design, PAGaS can be applied on top of any reconstruction method to enhance its high-frecuency details even boosting its global metrics. Qualitative results are especially telling, showing that gains come from enhancing fine-grained depth and mesh details, which is the core goal of our method.

\subsection{Qualitative Results}
\cref{fig:normals_and_meshes_pgsr} presents normals and 3D meshes from PGSR before and after refinement with PAGaS across scenes of increasing scale, from compact indoor settings to large outdoor environments. The refined normal maps provide the clearest evidence of PAGaS’s ability to recover high-frequency, pixel-level surface structure. Smaller scenes permit a finer TSDF voxel size, which better preserves these subtle refinements. In contrast, larger scenes require coarser voxels that inevitably smooth out fine geometric variations. This explains the reduced performance gap observed in the outdoor evaluations of \cref{tab:metrics_tnt}, particularly in large scenes such as Courthouse and Meetingroom, versus the clear improvements in the small object scenes reported in \cref{tab:metrics_dtu}.
Interestingly, all examples in \cref{fig:normals_and_meshes_pgsr} correspond to cases where the PAGaS-refined mesh is quantitatively worse than the strongest baseline PGSR, despite visibly exhibiting more reliable local detail in the DTU examples and to a lesser extent in Barn. This highlights that conventional geometric metrics can be misleading, especially on large-scale scenes, when evaluating methods like PAGaS whose primary contribution is enhancing fine-grained, local surface depth details. 
PAGaS reaches its full potential in consistently illuminated, object-centric, high resolution, dense multi-view capture setups.

The ActorsHQ dataset embodies all these characteristics, making it an ideal setting for testing this technology. As shown in \cref{fig:actor_normal_and_mesh}, PAGaS captures extremely fine geometric detail while scaling efficiently and keeping computational requirements within practical limits.

Extensive additional qualitative results of the refined depths and meshes from MVSAnywhere, 2DGS and PGSR in DTU and TnT, and for COLMAP in BlendedMVS, are documented in the supplementary.

\section{Conclusion}
\label{sec:conclusion}

We introduce Pixel-Aligned 1DoF Gaussian Splatting (PAGaS), a new splatting method tailored for multi-view depth refinement. 
Our key innovation lies in a minimal 1DoF Gaussian parametrization: whereas 3DGS optimizes 59 parameters per Gaussian, PAGaS optimizes only a single one. 
Gaussian in PAGaS are set opaque, of fixed color, spherical, pixel-aligned and constrained to move and scale exclusively along its back-projected ray, which leaves pixel-wise depth the sole optimizable variable. 
This radical simplification yields a method that is faster, memory-efficient, and capable of producing highly detailed results. 
We further introduce the Occlusion-Aware 3DGS Rasterizer, which adapts the alpha-blending logic of 3DGS to occlude Gaussians belonging to different surfaces, avoiding additional computation to increase or modify the existing Gaussians.

To the best of our knowledge, PAGaS is the first multi-view stereo method that leverages full-resolution input images to predict high fidelity depth at matching resolution, all while remaining scalable, requiring no pretraining, avoiding excessive memory use, and running within practical time budgets. Across DTU, TnT, ActorsHQ and BlendedMVS, PAGaS consistently refines depths and meshes, capturing fine-grained high-frequency details that both geometric and learning-based MVS methods typically miss.

We see PAGaS as a practical, high-precision post-processing module for existing reconstruction pipelines and a strong basis for future work in high-fidelity depth estimation under realistic compute and memory constraints. Rather than altering global geometry, its goal is to recover the subtle high-frequency details present in the original high-resolution imagery, details that are lost when downsampled, resulting in accurate and visually compelling refinements of coarse reconstructions.
{
    \small
    \bibliographystyle{ieeenat_fullname}
    \bibliography{main}
}
\clearpage
\setcounter{page}{1}
\setcounter{table}{0}
\setcounter{figure}{0}

\maketitlesupplementary
\appendix

\startcontents[supp]


\counterwithin{table}{section}
\counterwithin{figure}{section}

\renewcommand{\thetable}{S\thesection.\arabic{table}}
\renewcommand{\thefigure}{S\thesection.\arabic{figure}}

\section{Visual comparison with the ground truth}

\begin{figure}[ht!]
  \centering
  \includegraphics[width=.4\textwidth, keepaspectratio]{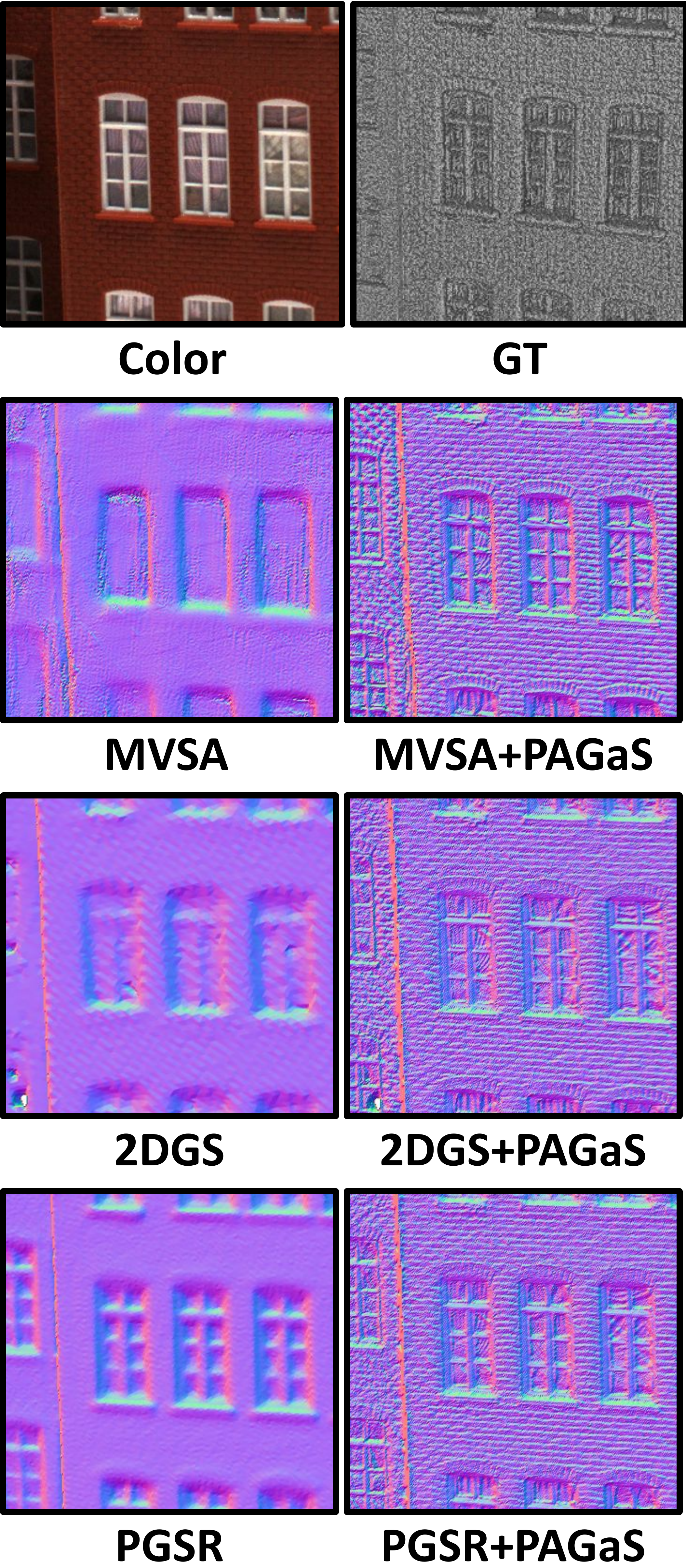}
  \caption{
  \textbf{Comparison of the normals} from depth for the baselines MVSAnywhere, 2DGS, and PGSR \textbf{before and after refinement} with our PAGaS, and the color \textbf{and the ground truth point cloud} in DTU scan24. See how PAGaS adds state-of-the-art high-frequency surface details independently of the quality of the initialization method. Notice that the individual building blocks are present in both the color and the ground truth, and PAGaS is the only method currently capable of capturing this level of detail.
  }
  \label{fig:dtu_gt}
\end{figure}

~\cref{fig:dtu_gt} extends the teaser in~\cref{fig:teaser} by providing results for the baselines (MVSAnywhere, 2DGS, and PGSR) and additionally visualizing the color and ground truth point cloud for DTU scan24. PAGaS consistently restores high-frequency geometric details that are lost in the baselines. It is the only method that accurately recovers the fine building-block structure visible in both the color and the ground truth. Full images and meshes can be found in \cref{fig:normals_scan24}. There, it can be observed that the fine details added in the per-view refined depth maps are preserved after fusion into the global 3D meshes, demonstrating that our method does not introduce locally inconsistent details but instead produces globally consistent ones.

\section{TnT inconsistent camera exposure problem}
\label{sec:exposure}

\begin{figure}[hb!]
  \centering
  \includegraphics[width=.5\textwidth, keepaspectratio]{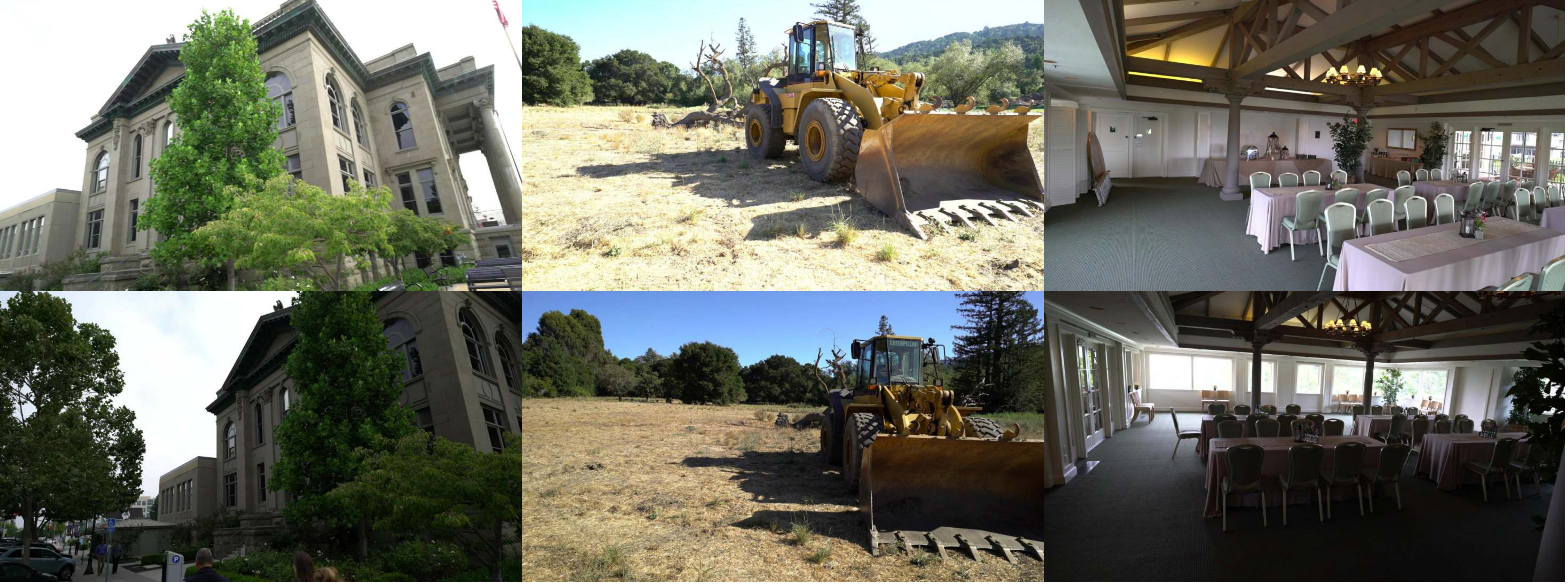}
  \caption{\textbf{Inconsistent camera exposure problem on TnT}. Each image was captured with a different and unknown camera exposure. This misleads photometric optimization methods that assume pixel color consistency across stereo views to triangulate 3D points, including 2DGS and our PAGaS.}
  \label{fig:tnt_illumination}
\end{figure}

\begin{table*}[t]
    \centering
    \footnotesize
    \setlength{\tabcolsep}{4pt}
    \resizebox{.8\textwidth}{!}{%
    \begin{tabular}{lccccccc}
        \toprule
        \textbf{Scene} & \textbf{Barn} & \textbf{Caterpillar} & \textbf{Courthouse} & \textbf{Ignatius} & \textbf{Meetingroom} & \textbf{Truck} & \textbf{Mean} \\
        \midrule        
        MVSAnywhere + PAGaS w/o exposure
            & 0.29 & 0.12 & 0.09 & 0.25 & 0.15 & 0.21 & 0.18 \\
        MVSAnywhere + PAGaS w/ exposure
            & \nd0.32 & \nd0.14 & \nd0.09 & \nd0.28 & \nd0.16 & \nd0.22 & \st0.20 \\
        \noalign{\vskip 1mm} \hdashline \noalign{\vskip 1mm}
        2DGS + PAGaS w/o exposure
            & 0.42 & 0.27 & 0.17 & 0.57 & \nd0.05 & 0.19 & 0.28 \\
        2DGS + PAGaS w/ exposure
            & \nd0.47 & \nd0.33 & \nd0.22 & \nd0.68 & 0.04 & \nd0.19 & \st0.32 \\
        \noalign{\vskip 1mm} \hdashline \noalign{\vskip 1mm}
        PGSR + PAGaS w/o exposure
            & 0.49 & 0.40 & 0.21 & 0.75 & 0.13 & \nd0.23 & 0.37 \\
        PGSR + PAGaS w/ exposure
            & \nd0.52 & \nd0.45 & \nd0.24 & \nd0.81 & \nd0.14 & \nd0.23 & \st0.40 \\
        \bottomrule
    \end{tabular}}
    \caption{\textbf{Quantitative TnT evaluation with and without camera exposure compensation} showing the F1-score $\uparrow$ for baseline methods before and after our PAGaS refinement.}
    \label{tab:ablations_tnt}
\end{table*}

The DTU dataset was captured in a controlled studio environment with stable camera exposures. In contrast, TnT contains frames with unknown and varying exposures, as shown in \cref{fig:tnt_illumination}. Purely photometric optimization methods whose only input is the set of color images, and which rely on no external priors to constrain the problem, are therefore highly sensitive to the quality of these images. Large exposure shifts between adjacent views make the downstream reconstruction task more difficult, leading to degraded results.
PGSR explicitly optimizes a per-image illumination and exposure model, which reduces the impact of these variations. Motivated by this observation, in \cref{tab:ablations_tnt} we evaluate the effect of using a similar exposure model and a multi-view stereo consistency check during our PAGaS refinements on top of all baselines.
Since neither 2DGS nor MVSAnywhere compensate for exposure in the input images, we also avoid compensating for them when refining their depths with PAGaS in the main experiments in \cref{tab:metrics_tnt}. This allows us to demonstrate, under identical input conditions, that PAGaS consistently improves their reconstructions.

\section{Ablations}

\begin{table}[htb!]
    \centering
    \footnotesize
    \setlength{\tabcolsep}{1pt}
    \newcommand{\colsep}{\hspace{16pt}}
    \resizebox{.5\textwidth}{!} {%
    \begin{tabular}{@{\extracolsep{4pt}} c c @{\colsep} c c @{\extracolsep{4pt}}}%
        \toprule
        \textbf{2DGS} & \textbf{2DGS + PAGaS} & \textbf{2DGS} & \textbf{2DGS + PAGaS}  \\         
        \midrule
        w/o Depth from mesh & 0.564 & 200, 100 Steps & 0.468 \\
        Random depth & 10,79 & 100, 200 Steps & 0.458   \\ 
        1 Context view & 0.465 & w/o Fix opacity & 0.457   \\     
        2 Context views & 0.464 & w/o Scale from depth & 0.479 \\    
        3 Context views &0.464 & Learning rate 1e-4 & 0.563  \\   
        4 Context views & 0.463 & 3 Radius threshold & 0.461  \\       
        5 Context views & 0.462 & 4 Depth Slices & 0.460    \\  
        6 Context views & 0.461 & w/o Occlusion-Aware Rasterizer & 0.464   \\  
        7 Context views & 0.461 & w/o Smoothness & 0.457 \\  
        8 Context views & 0.459 & View-dependent color & 0.454 \\
        9 Context views &0.458 & 0,8 $\lambda_c$ & 0.456  \\  
        w/o Scales & 0.460 & w/ Exposure compensation & 0.458  \\  
        4, 2, 1 Scales & 0.495 & w/ Stereo consistency & 0.471 \\  
        8, 4, 2, 1 Scales &0.630 & Full model & 0.456 \\  
        \bottomrule
    \end{tabular}}
    \caption{\textbf{Ablation study} of the characteristic parameters of our PAGaS. Showing Chamfer distance in mm (lower better) for 2DGS after PAGaS refinement in scan24 of DTU dataset.} 
    \label{tab:ablations}
\end{table}

\cref{tab:ablations} summarizes the ablations of the key parameters of PAGaS. 

If instead of using the depth extracted from the baseline mesh, we directly input the baseline’s estimated depth for that view, this leads to poorer convergence. PAGaS is a refinement method: it adds small-scale surface detail to the input depth without changing its global structure. The more stereo-consistent the input depth is, the better PAGaS performs, because the details refined in one view align with those captured in the others.

If no initialization depth is provided (i.e., we start from random depth bounded by the cameras' pose range), PAGaS fails to converge. Its role is to move the Gaussians attached to each pixel’s depth slightly along the ray, not to correct large errors or traverse long distances.

Accuracy improves as more context views are used to constrain the optimization. Notably, even a small number of views already approaches the full-model performance, demonstrating the robustness and stability of our method. This also means the number of context views can be reduced to save GPU memory and increase speed with little loss in quality. Our full model with 10 context views at full resolution uses 8 GB of VRAM, whereas using a single context view requires only 2 GB and runs about 4× faster.

We use a two-scale pyramid in all experiments, since the baseline depth is typically produced at half the image resolution. Using more scales (i.e., more aggressive initial downscaling) makes the starting depth too coarse, whereas PAGaS benefits from an initialization that is already a close approximation to the true scene geometry.

Another experiment to highlight is modeling per-Gaussian view-dependent color using standard 3rd-order spherical harmonics. This increases the per-Gaussian parameter count from 1 (depth) to 48 (3 RGB channels×16 SH coefficients). Accuracy remains essentially unchanged, but GPU VRAM usage grows substantially (it no longer fits on an 11 GB GPU) and the optimization takes roughly twice as long.

As explained in \cref{sec:exposure}, the DTU dataset has consistent camera exposure across all images. Consequently, applying per-view exposure compensation or a multi-view stereo consistency check on DTU does not provide any benefit, as the photometric optimization remains stable across views.

\section{Depth vs normal visualization}

\begin{figure}[ht!]
  \centering
  \includegraphics[width=.55\textwidth, keepaspectratio]{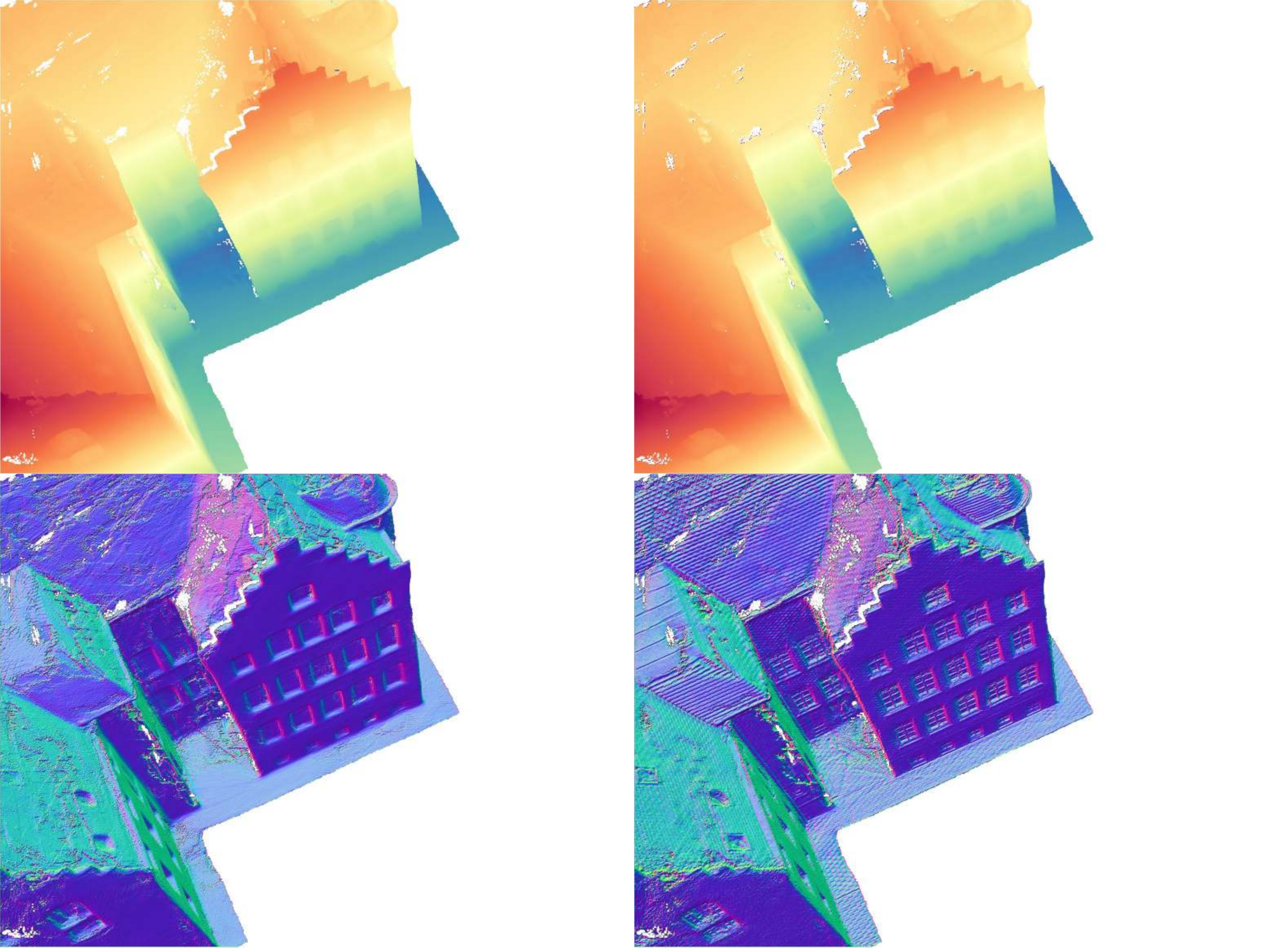}
  \caption{
  \textbf{Depth and normal comparison on DTU scan24.} Left: depth and normals from the original MVSAnywhere mesh. Right: results after applying our PAGaS refinement. Although depth maps may appear similar at first glance, the normals highlight the substantial improvement in fine-scale surface geometry.
  }
  \label{fig:depth_vs_normal}
\end{figure}

\cref{fig:depth_vs_normal} illustrates the effect of our refinement by comparing the depth and corresponding normal maps on DTU scan24 before and after applying PAGaS. While depth maps alone tend to obscure fine-scale geometry due to the small magnitude of relative depth variations, the normal maps clearly reveal the real quality of the baselines and the improvements introduced by our method. As shown, PAGaS enhances high-frequency surface structure that is largely absent in the original MVSAnywhere reconstruction.

\section{Additional qualitative evaluations}

\begin{figure*}[ht!]
  \centering
  \includegraphics[height=.95\textheight, keepaspectratio]{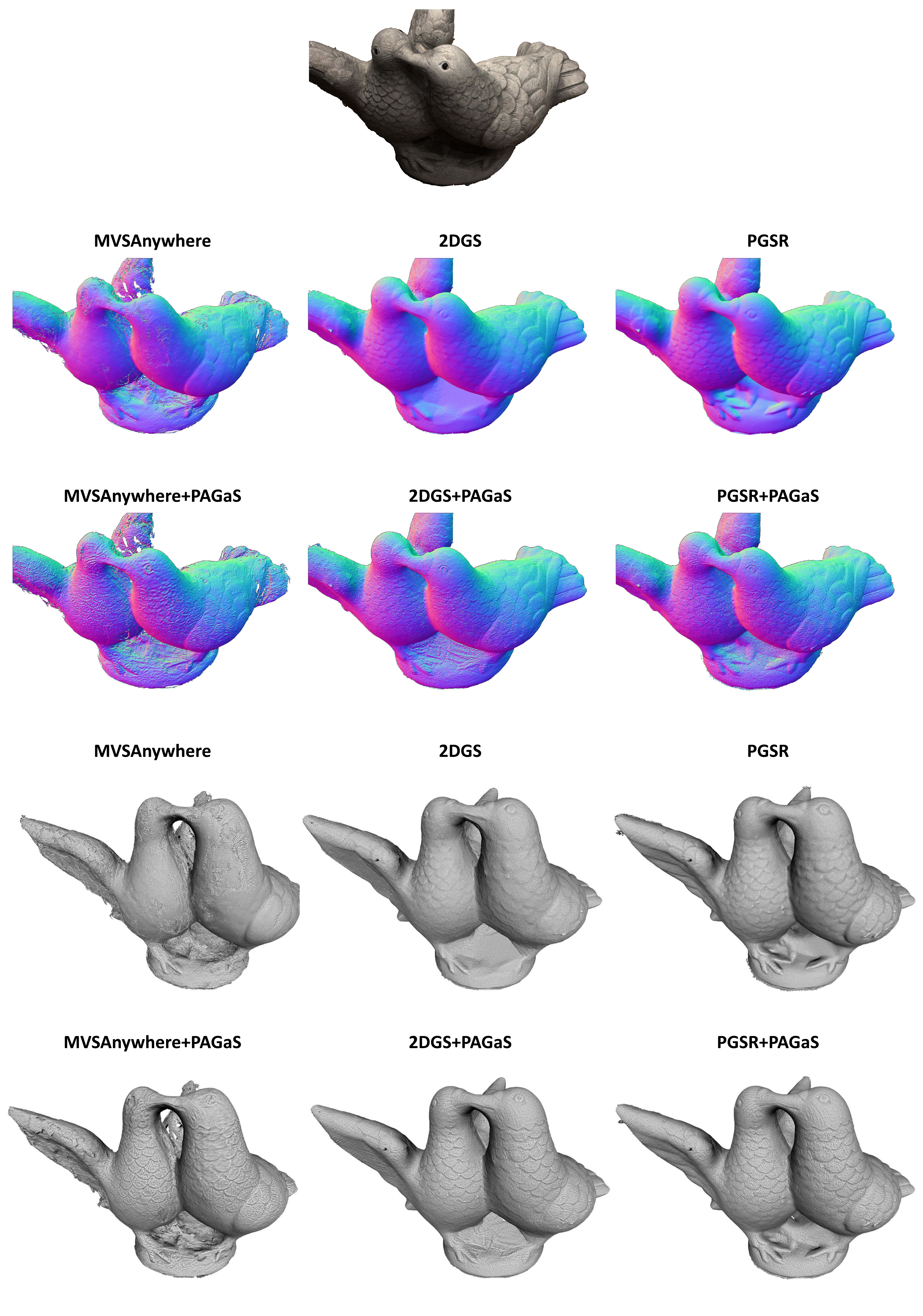}
  \caption{\textbf{Qualitative evaluation of the normals from depth and the 3D meshes in DTU scan106} for the three main baselines before and after refinement with our PAGaS. Zoom in to appreciate the added pixel-wise details.}
  \label{fig:normals_scan106}
\end{figure*}

\begin{figure*}[ht!]
  \centering
  \includegraphics[height=.95\textheight, keepaspectratio]{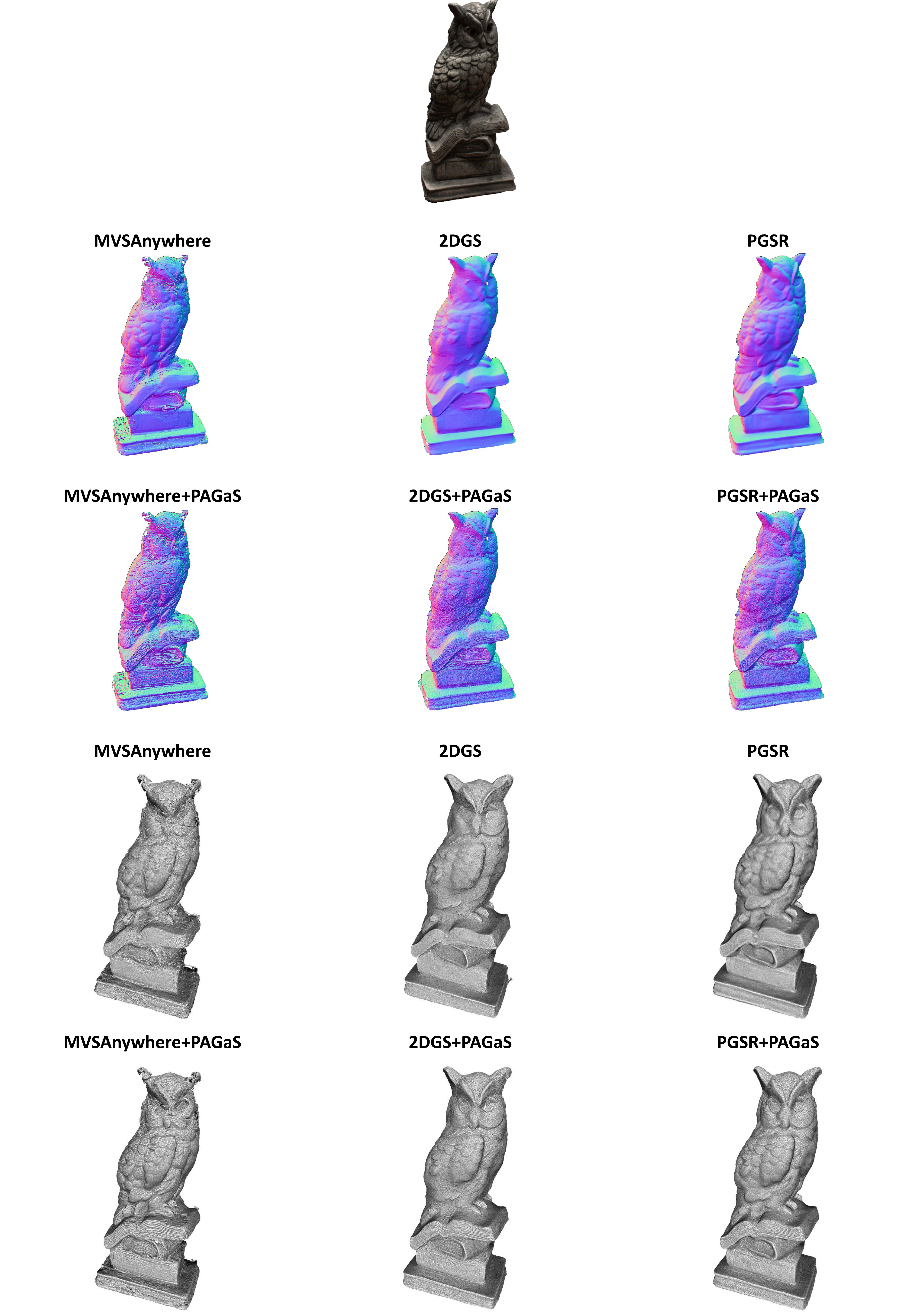}
  \caption{\textbf{Qualitative evaluation of the normals from depth and the 3D meshes in DTU scan122} for the three main baselines before and after refinement with our PAGaS. Zoom in to appreciate the added pixel-wise details.}
  \label{fig:normals_scan122}
\end{figure*}

\begin{figure*}[ht!]
  \centering
  \includegraphics[height=.95\textheight, keepaspectratio]{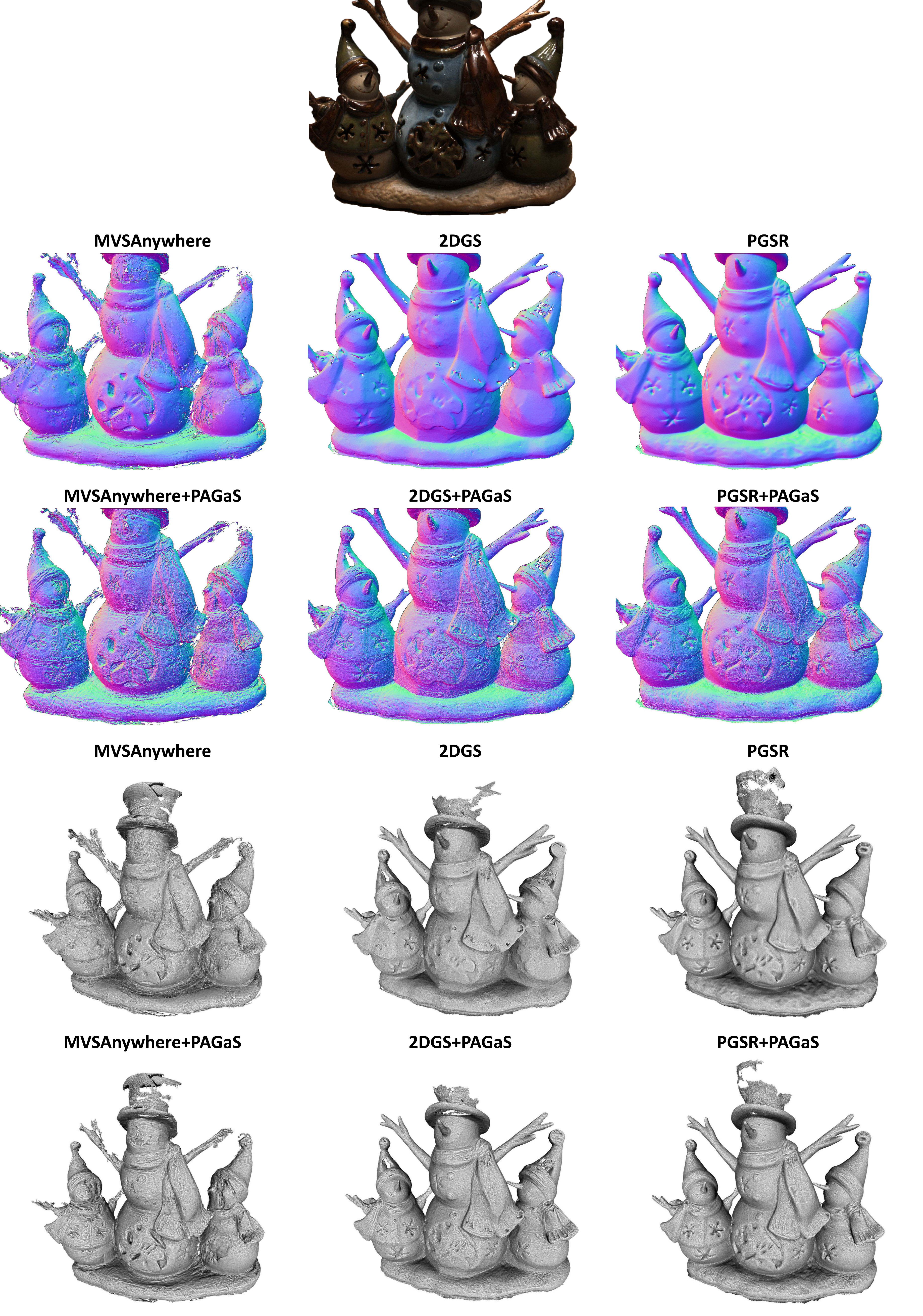}
  \caption{\textbf{Qualitative evaluation of the normals from depth and the 3D meshes in DTU scan69} for the three main baselines before and after refinement with our PAGaS. Zoom in to appreciate the added pixel-wise details.}
  \label{fig:normals_scan69}
\end{figure*}

\begin{figure*}[ht!]
  \centering
  \includegraphics[height=.95\textheight, keepaspectratio]{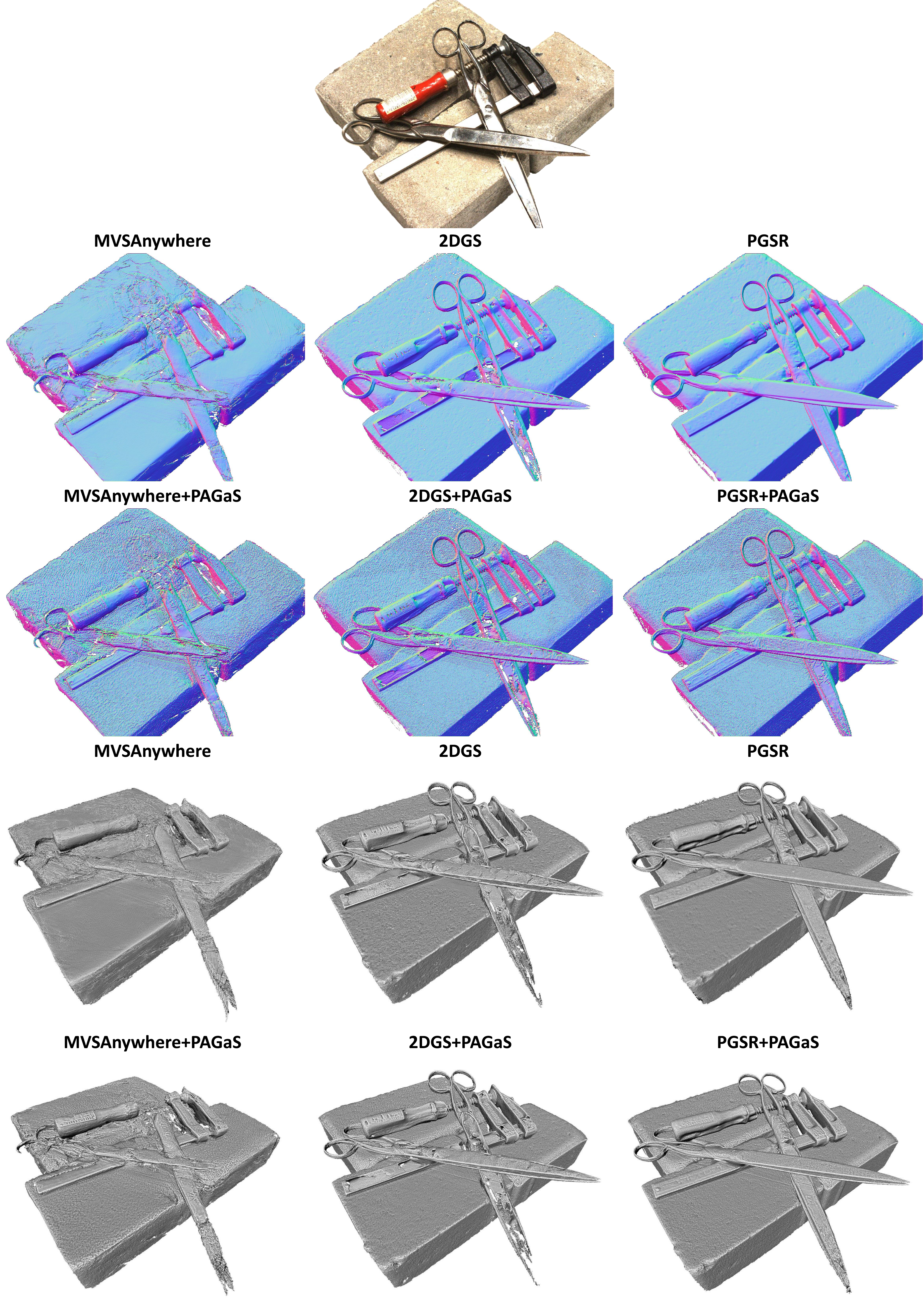}
  \caption{\textbf{Qualitative evaluation of the normals from depth and the 3D meshes in DTU scan37} for the three main baselines before and after refinement with our PAGaS. Zoom in to appreciate the added pixel-wise details.}
  \label{fig:normals_scan37}
\end{figure*}

\begin{figure*}[ht!]
  \centering
  \includegraphics[height=.95\textheight, keepaspectratio]{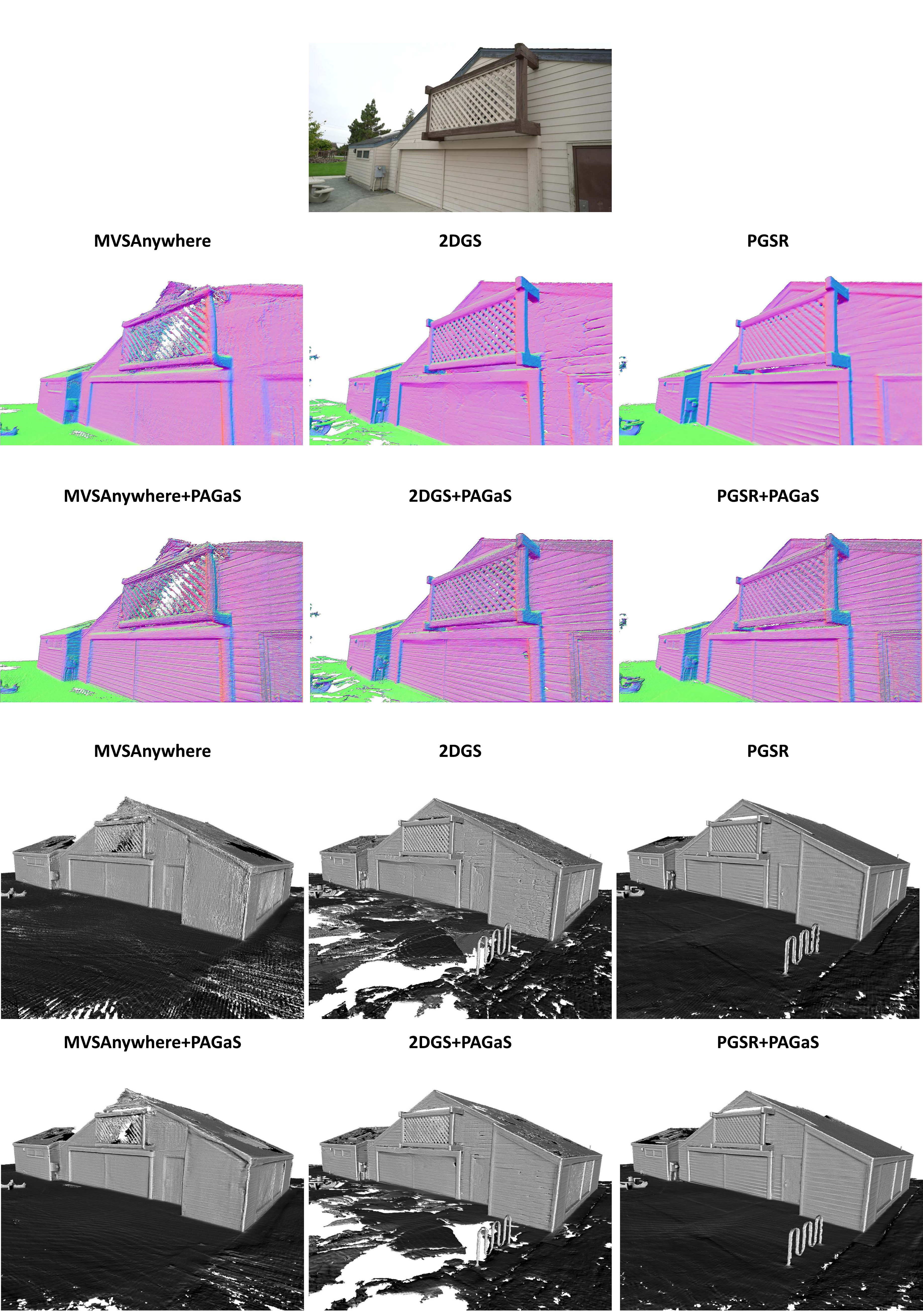}
  \caption{\textbf{Qualitative evaluation of the normals from depth and the 3D meshes in TNT Barn} for the three main baselines before and after refinement with our PAGaS. Zoom in to appreciate the added pixel-wise details.}
  \label{fig:normals_barn}
\end{figure*}

\begin{figure*}[ht!]
  \centering
  \includegraphics[height=.95\textheight, keepaspectratio]{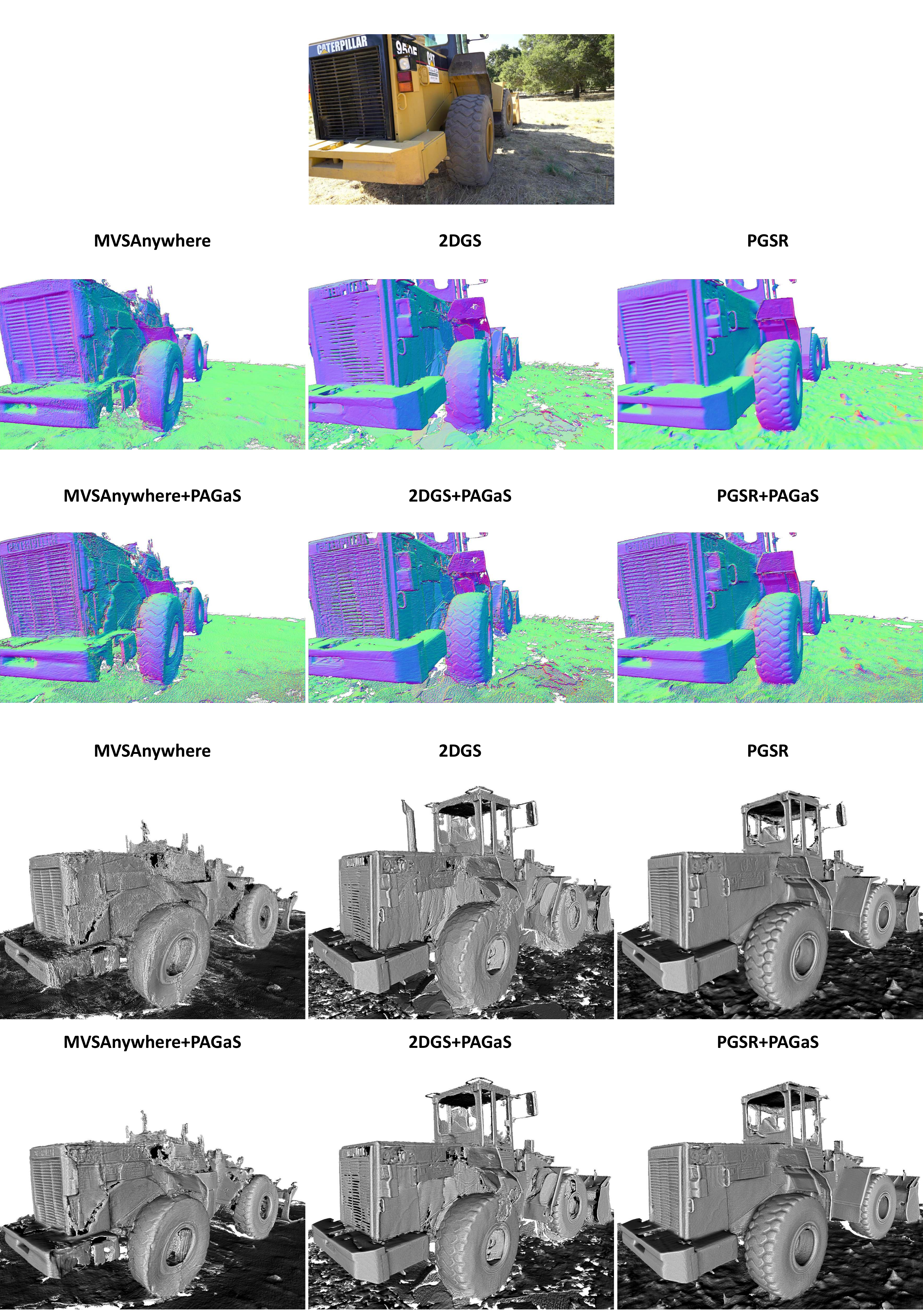}
  \caption{\textbf{Qualitative evaluation of the normals from depth and the 3D meshes in TNT Caterpillar} for the three main baselines before and after refinement with our PAGaS. Zoom in to appreciate the added pixel-wise details.}
  \label{fig:normals_caterpillar}
\end{figure*}

\begin{figure*}[ht!]
  \centering
  \includegraphics[height=.95\textheight, keepaspectratio]{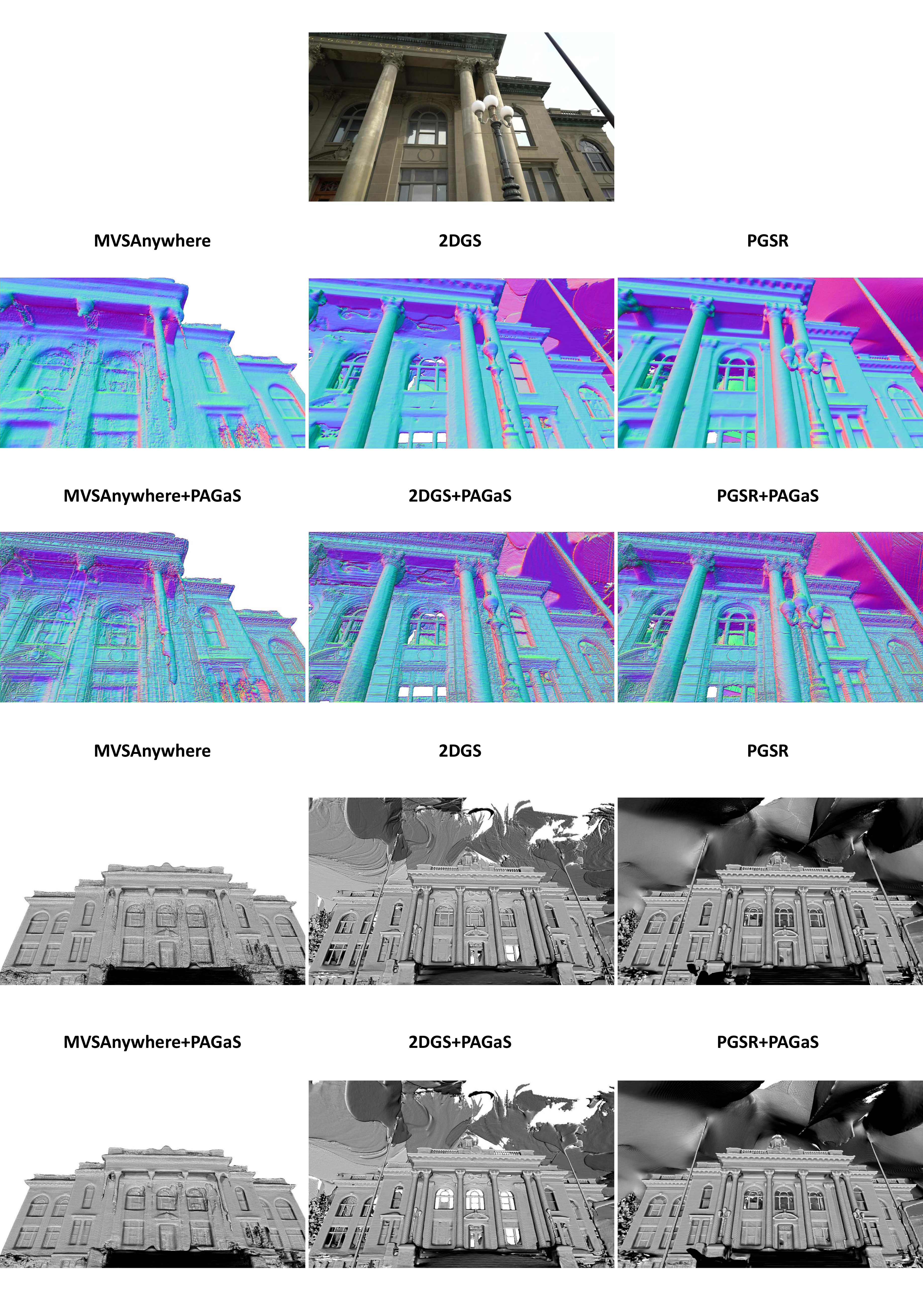}
  \caption{\textbf{Qualitative evaluation of the normals from depth and the 3D meshes in TNT Courthouse} for the three main baselines before and after refinement with our PAGaS. Zoom in to appreciate the added pixel-wise details.}
  \label{fig:normals_courthouse}
\end{figure*}

\begin{figure*}[ht!]
  \centering
  \includegraphics[height=.95\textheight, keepaspectratio]{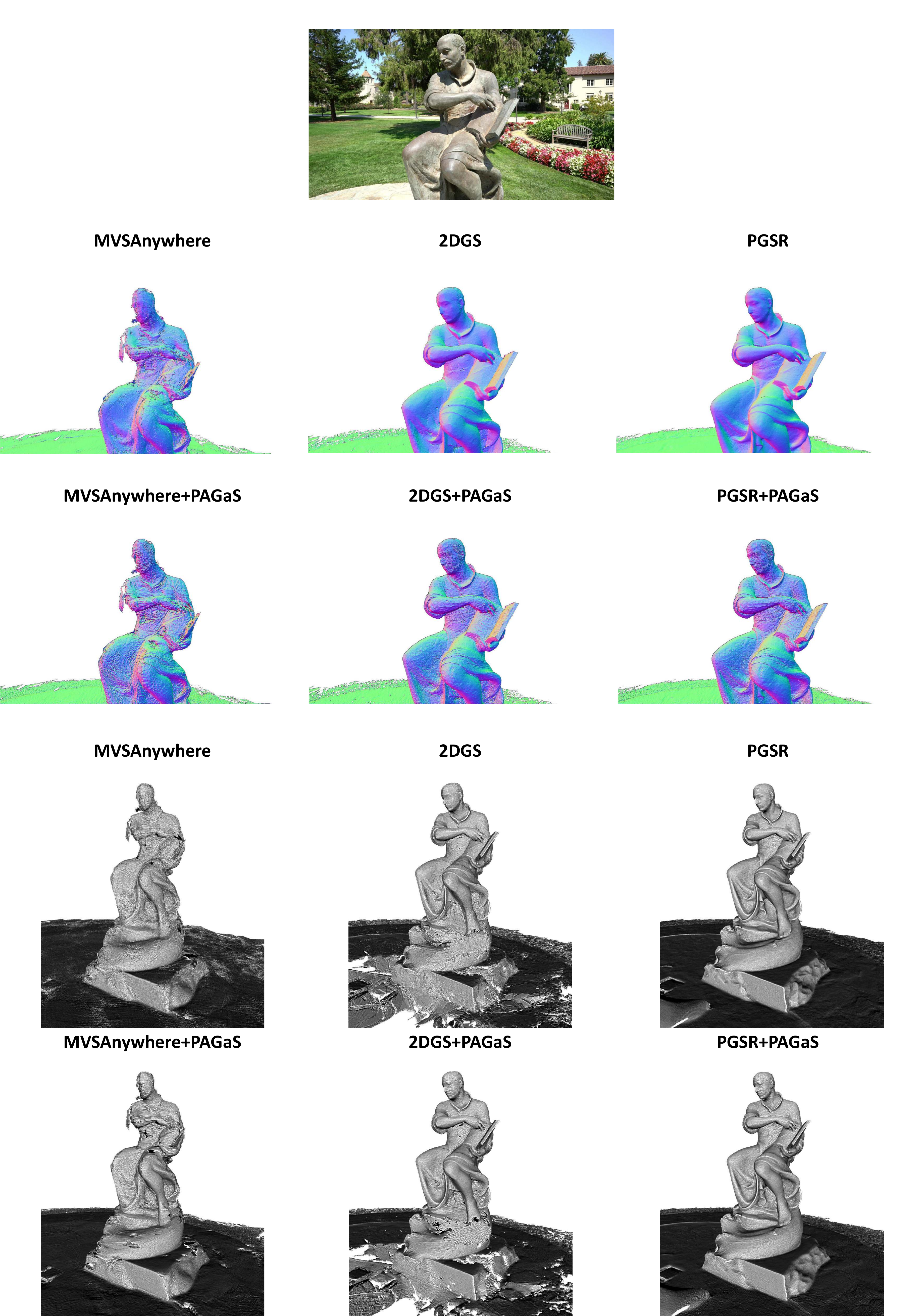}
  \caption{\textbf{Qualitative evaluation of the normals from depth and the 3D meshes in TNT Ignatius} for the three main baselines before and after refinement with our PAGaS. Zoom in to appreciate the added pixel-wise details.}
  \label{fig:normals_ignatius}
\end{figure*}

\begin{figure*}[ht!]
  \centering
  \includegraphics[height=.95\textheight, keepaspectratio]{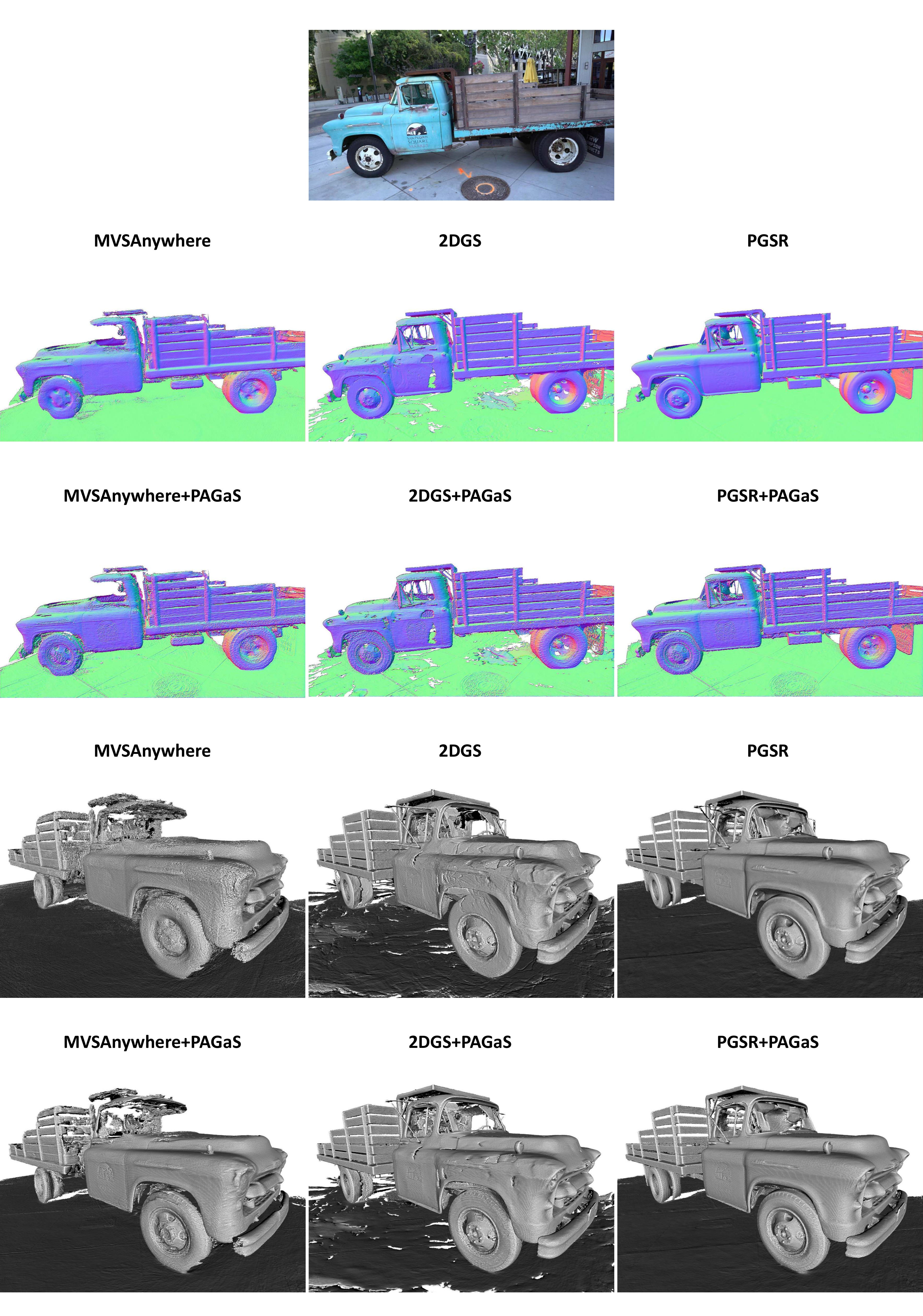}
  \caption{\textbf{Qualitative evaluation of the normals from depth and the 3D meshes in TNT Truck} for the three main baselines before and after refinement with our PAGaS. Zoom in to appreciate the added pixel-wise details.}
  \label{fig:normals_truck}
\end{figure*}

\begin{figure*}[ht!]
  \centering
  \includegraphics[height=.95\textheight, keepaspectratio]{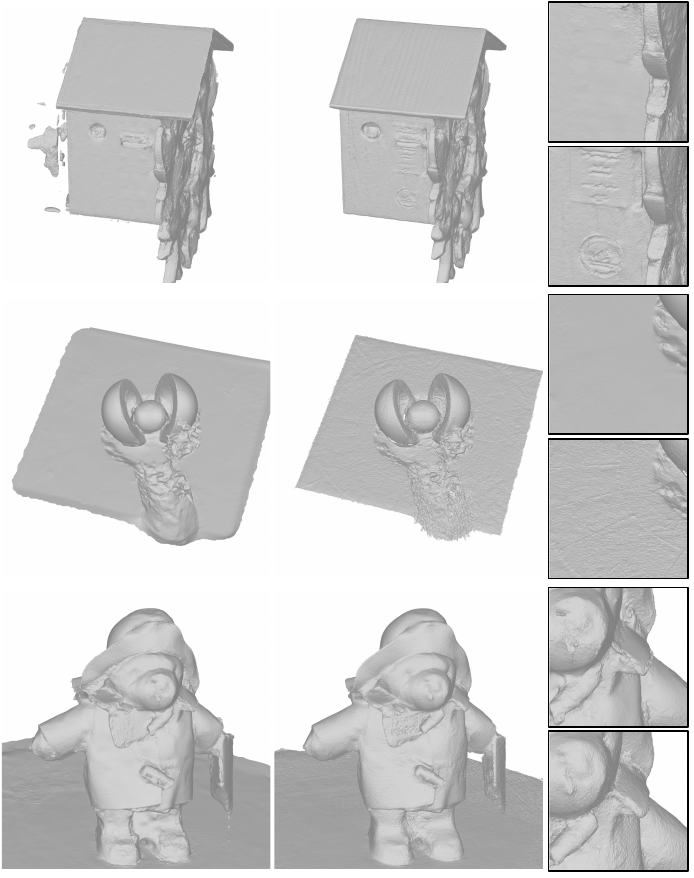}
  \caption{\textbf{BlendedMVS meshes} by COLMAP's multi-view stereo before (left) and after (right) applying our PAGaS with some close-up areas. From top to bottom: clock, stone, bear. Zoom in to observe the fine-grained details refined by our PAGaS.}
  \label{fig:bmvs}
\end{figure*}

Extensive additional qualitative evaluations of the baselines MVSAnywhere, 2DGS and PGSR before and after our PAGaS refinement are provided for DTU in scans 24, 106, 122, 69 and 37 at~\cref{fig:normals_scan24,fig:normals_scan106,fig:normals_scan122,fig:normals_scan69,fig:normals_scan37}; for TnT in Barn, Caterpillar, Courthouse, Ignatius and Truck at~\cref{fig:normals_barn,fig:normals_caterpillar,fig:normals_courthouse,fig:normals_ignatius,fig:normals_truck}; and of Colmap for BlendedMVS at~\cref{fig:bmvs}.

\begin{figure*}[ht!]
  \centering
  \includegraphics[height=.95\textheight, keepaspectratio]{figs/normals_scan24.pdf}
  \caption{\textbf{Qualitative evaluation of the normals from depth and the 3D meshes in DTU scan24} for the three main baselines before and after refinement with our PAGaS. Zoom in to appreciate the added pixel-wise details.}
  \label{fig:normals_scan24}
\end{figure*}

\end{document}